%% file: main.tex
\documentclass{article}

\usepackage{microtype}
\usepackage{graphicx}
\usepackage{subfigure}
\usepackage{booktabs} % for professional tables

\usepackage{hyperref}
\usepackage[normalem]{ulem}
\usepackage{amsmath}
\usepackage[dvipsnames]{xcolor}
\usepackage{soul}
\usepackage{multirow}
\usepackage{mdframed}
\usepackage{cuted}
\usepackage[most]{tcolorbox}
\tcbuselibrary{breakable}
\newmdenv[linecolor=black, linewidth=1pt, skipabove=\topsep, skipbelow=\topsep]{myframedbox}

\newcommand{\delete}[1]{\{\textit{\sout{#1}}\}}
\newcommand{\sdelete}[1]{\textit{\sout{#1}}}
\newcommand{\colorize}[2]{\colorbox{Salmon!#1!white}{\strut #2}}

\usepackage[accepted]{icml2023}

\icmltitlerunning{Selective Context: Enhancing Context Efficiency of LLMs}

\begin{document}

\twocolumn[
% \icmltitle{\textit{Selective Context}: Enhancing Context Efficiency of LLMs with Self-Information-Based Content Filtering}
\icmltitle{Unlocking Context Constraints of LLMs: Enhancing Context Efficiency of LLMs with Self-Information-Based Content Filtering }




\begin{icmlauthorlist}
\icmlauthor{Yucheng Li}{U}
% \icmlauthor{Aeiau Zzzz}{equal,to}
% \icmlauthor{Bauiu C.~Yyyy}{equal,to,goo}
% \icmlauthor{Cieua Vvvvv}{goo}
% \icmlauthor{Iaesut Saoeu}{ed}
% \icmlauthor{Fiuea Rrrr}{to}
% \icmlauthor{Tateu H.~Yasehe}{ed,to,goo}
% \icmlauthor{Aaoeu Iasoh}{goo}
% \icmlauthor{Buiui Eueu}{ed}
% \icmlauthor{Aeuia Zzzz}{ed}
% \icmlauthor{Bieea C.~Yyyy}{to,goo}
% \icmlauthor{Teoau Xxxx}{ed}
% \icmlauthor{Eee Pppp}{ed}
\end{icmlauthorlist}

\icmlaffiliation{U}{University of Surrey, UK}

% \icmlaffiliation{to}{Department of Computation, University of Torontoland, Torontoland, Canada}
% \icmlaffiliation{goo}{Googol ShallowMind, New London, Michigan, USA}
% \icmlaffiliation{ed}{School of Computation, University of Edenborrow, Edenborrow, United Kingdom}

\icmlcorrespondingauthor{Yucheng Li}{yucheng.li@surrey.ac.uk}
% \icmlcorrespondingauthor{Eee Pppp}{ep@eden.co.uk}

\icmlkeywords{Machine Learning, ICML}

\vskip 0.3in
]



\printAffiliationsAndNotice{}  % leave blank if no need to mention equal contribution

\begin{abstract}
Large language models (LLMs) have received significant attention by achieving remarkable performance across various tasks. However, their fixed context length poses challenges when processing long documents or maintaining extended conversations. This paper proposes a method called \textit{Selective Context} that employs self-information to filter out less informative content, thereby enhancing the efficiency of the fixed context length. We demonstrate the effectiveness of our approach on tasks of summarisation and question answering across different data sources, including academic papers, news articles, and conversation transcripts.
\end{abstract}

\section{Introduction}

Large language models (LLMs) have demonstrated remarkable power and impressive generalisation abilities across a wide range of natural language processing tasks, as well as real-life applications \cite{brown2020language,touvron2023llama,bubeck2023sparks}. However, a major limitation of LLMs is their fixed context length. As LLMs have no memory outside their context window, it poses a significant challenge when tackling tasks that involve processing long documents or engaging in extended conversations \cite{dong2023survey}. Increasing the context length for LLMs, particularly those based on Transformer, is very expensive due to the quadratic growth of memory and computation associated with the 2-D attention matrix \cite{vaswani2017attention}. These limitations highlight the need for more efficient solutions to utilize the limited context in tasks that require extended context.

Fortunately our experiments reveal that LLMs do not need all content in a document or the entire conversation history to answer users' queries. As shown in Figure \ref{fig:example}, LLMs are able to generate the expected answer even with relevant information deleted. This might be because LLMs can infer the missing information based on the contextual clues and prior knowledge acquired from their pre-training. As a result, we argue that optimizing the use of context length by filtering out less informative content is possible without sacrificing performance.

In this paper, we propose \textit{Selective Context}, which filters out less informative content to reduce the cost of a given context, thereby making better use of the fixed context length in LLMs. \textit{Selective Context} employs a base language model to compute self-information for lexical units (sentences, phrases, or tokens) in a context and use it to evaluate their informativeness. By selectively retaining content with higher self-information, our method provides a more compact and efficient context representation for LLMs to process without compromising their performance on various tasks.

\begin{figure}[t]
    \centering
    \caption{LLMs are able to answer correctly with less informative content deleted.}
    \label{fig:example}

    \vskip 0.1in
    
    \fbox{\parbox[c]{\columnwidth}{
        \textbf{Context: } Large Languages Models (LLMs) \delete{trained on massive corpora of texts} have shown their ability to perform new tasks from textual instructions or from a few examples, which resulting in a line of work that focuses on further scaling these models. These efforts are based on the assumption \delete{that more parameters will lead to better performance.}

        \vskip 0.1in

        \textbf{Query: } What's the assumption behind?

        \vskip 0.1in

        \textbf{GPT: } Further scaling Large Language Models will lead to better performance on a wide range of tasks.
    }}
\end{figure}

To evaluate the effectiveness of our proposed method, we tested Selective Context on three data sources: arxiv papers, BBC news articles, and conversation transcripts with four different NLP tasks: summarisation, question answering, original context reconstruction and conversation. Our results demonstrate that Selective Context significantly enhances the efficiency of LLMs, allowing them to handle long documents and extended conversations with only minor sacrifices in generation quality.

The key contributions of our paper: (1) We introduce Selective Context, a novel approach to context filtering that maximises the utility of fixed context length in LLMs. (2) We provide extensive evaluations of the proposed method. (3) Our results demonstrate the effectiveness of Selective Context in reducing the cost of context in LLMs. Code and data can be found in \url{https://github.com/liyucheng09/Selective_Context}.

\section{Self-Information}

Self-information, also known as \textit{surprisal} or \textit{information content}, is a fundamental concept in information theory that quantifies the amount of information conveyed by an event \cite{shannon1948mathematical}. In the context of language modelling, the event here is one step of generation (i.e., a token). It is defined as the negative log likelihood of the token:

\begin{equation}
I(x) = -\log_2{P(x_t | x_{0}, x_{1}, ... ,x_{t-1})}
\end{equation}

where $I(x)$ represents the self-information of token $x$ and $P(x)$ denotes its output probability. 

In the information theory, self-information measures the level of surprise or uncertainty associated with an event; rare events convey more information and thus have higher self-information, while common events convey less information and have lower self-information. In the context of language modelling, self-information can be used to assess the informativeness of lexical units, e.g., words, phrases, or sentences, to see which pieces of information are more likely to be novel or important for understanding the context.

Self-information is usually not directly used in NLP. Instead, terms closely related such as entropy and perplexity are widely used in language model optimisation and evaluation.
\begin{align}
& H(S) = \frac{1}{N}\Sigma_t I(x_t) \\
& PP(S) = 2^{H(S)}
\end{align}
where the entropy $H(S)$ of the sentence $S=(x_0, ..., x_n)$ is the average self-information of words in the sentence, and perplexity $PP(S)$ of the sentence can be calculated with entropy.

The property of self-information that is especially relevant to our method is the additivity.
\begin{align}
I(x_0, x_1) &= -\log_2{P(x_0, x_1)} \\
            &= -\log_2{P(x_0)P(x_1|x_0)} \\
            &= -\log_2{P(x_0)} - \log_2{P(x_1|x_0)} \\
            &= I(x_0)I(x_1)
\end{align}
It means we can calculate the self-information of a lexical unit by simply sum self-information of tokens in it.

\section{Method}

In this section, we present the details of our proposed method, \textit{Selective Context}, which optimizes the use of context length in LLMs by filtering out less informative content. The main idea is to compute the self-information for lexical units (such as sentences, phrases, or tokens) within a given context and utilise it to evaluate their informativeness. We first compute the self-information for each token in the context, then merge tokens and their self-information based on lexical units such as phrases or sentences. The overall approach consists of the following steps:

\subsection{Computing Self-Information}

Given a context $C = {x_0, x_1, ..., x_n}$, where $x_i$ denotes a token, we use a base language model $M$ to compute the self-information for each token $x_t$ as fellow:

\begin{equation}
I(x_i) = -\log_2{P(x_i | x_0, x_1, ..., x_{i-1})}
\end{equation}

The base language model here should be causal language model, such as GPT-2, OPT, and LLaMA. 

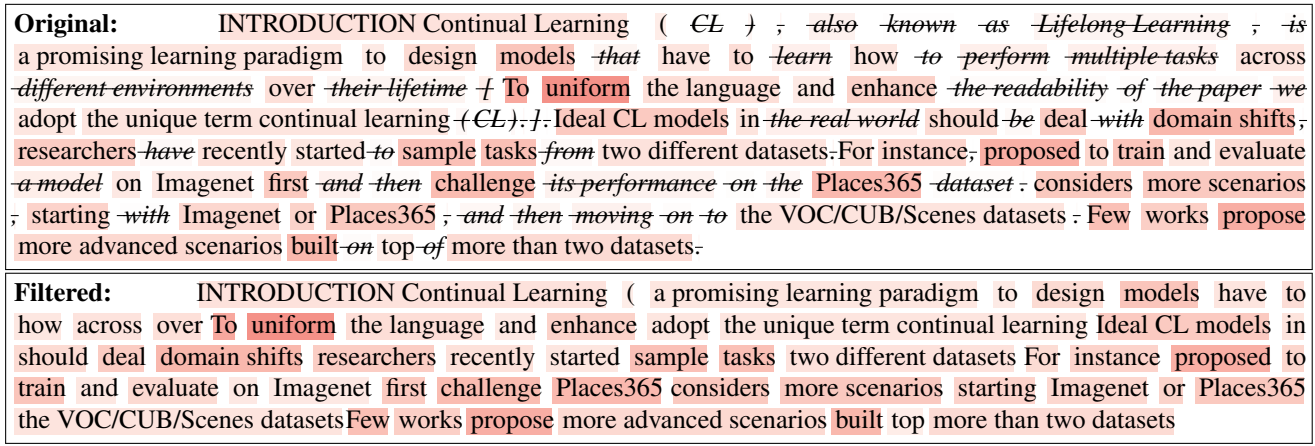
\begin{figure*}[t]
    \centering
    \fbox{\parbox[c]{\textwidth}{ 
\textbf{Original: }{\setlength{\fboxsep}{-1pt}
 \input{text_0}
}
   \par }}
\fbox{\parbox[c]{\textwidth}{
\textbf{Filtered: } {\setlength{\fboxsep}{-1pt}
 \input{text_1}
}
    }}
        \vskip -0.06in
    \caption{A visualisation of self-information based content filter. The paragraph is from a very recent paper.}
    \label{fig:color_text}
    \vskip -0.1in
    
\end{figure*}

\subsection{Merging into Lexical Units}

If the content filtering of selective context is directly performed on the token level, it might lead to very disjoint context. Therefore except token level filtering, we also conduct the filtering procedure in phrase and sentence level. We call a basic unit in our filtering a \textit{lexical unit}, which could be a token, a phrase or a sentence in our setting.

To enable selective context works on phrases and sentences, we should merge tokens and their self-information into lexical units. For each lexical unit $u = (x_t, ..., x_{t+\alpha})$, we can calculate its self-information by sum the self-information of its individual tokens according to the additivity property of self-information:

\begin{equation}
I(u) = \sum_{i=t}^{\alpha} I(x_i)
\end{equation}

Sentence tokenizer is employed to obtain sentence level lexical units. And we use spacy\footnote{\url{https://spacy.io/api/pipeline-functions\#merge_noun_chunks}} to merge tokens into nouns phrases. We do not merge verb phrases as it might produce super long phrases.

\subsection{Selective Retention of Informative Context}

With the self-information of each lexical unit computed, we can now evaluate their informativeness. Instead of using a fixed threshold or retaining a fixed number of top $k$ lexical units, we recommend to use a percentile-based filtering approach to adaptively select the most informative content.

First, we rank the lexical units based on their self-information values in descending order. Then, we compute the $p$-th percentile of self-information values among all lexical units.

\begin{equation}
I_{p} = \texttt{np.percentile}([I(u_0), .., I(u_k)], p)
\end{equation}

Next, we selectively retain lexical units with self-information values greater than or equal to the $p$-th percentile, constructing a filtered context $C'$:

\begin{equation}
C' = {U_i \mid I(U_i) \geq I_{p}, 1 \leq i \leq n}
\end{equation}

The percentile-based filtering is a more flexible approach to retain the most informative content depending on the distribution of self-information values in the given context. In Figure \ref{fig:color_text}, we present an example on phrase level where $p$ is set to 50, which means half of phrases are filtered out. In this case, the context after processed by selective context only remains 57.2\% tokens, which saves 42.7\% of context length. We will discussion how LLMs perform on the processed context in the next section.

\section{Experiments}

\subsection{Datasets}

We evaluate Selective Context on three datasets from different domains:

\noindent \textbf{BBC News:} A dataset containing news articles collected from the British Broadcasting Corporation (BBC) published in March 2023. This dataset covers a wide range of topics, including politics, business, sports, and technology. We use the full content of each news article in our experiments.

\noindent \textbf{Arxiv Articles:} A dataset consisting of latest academic papers created in March 2023 from the arXiv preprint repository. These papers span various scientific disciplines, such as computer science, physics, and mathematics. As Arxiv articles can be quite long, we only process the first two sections for each Arxiv paper in our experiments.

\noindent \textbf{Conversations from ShareGPT.com:} ShareGPT.com is a platform where ChatGPT users share their surprising and interesting conversation with ChatGPT. This datasets consists of conversations in different languages and in various scenarios (e.g., coding, chitchat, writing assistant, etc.). We use ShareGPT dataset for the conversation task in our experiments.

Statistics in detail are presented in Table \ref{tab:dataset_statistics}. Note that to avoid data contamination, we only collect latest data to ensure they are created after the knowledge cut off of ChatGPT. Data samples from the BBC News and Arxiv datasets were all created after March 2023. And conversations on SharedGPT.com are clearly created after the release of ChatGPT (\texttt{gpt-3.5-turbo}).

\begin{table}[t]
\centering
\caption{Statistics of the three datasets. \#Sent, \#Phrase, \#Token are averaged per document.}
\label{tab:dataset_statistics}
\vskip 0.1in
\begin{tabular}{lcccc}
\toprule
Dataset & \#Doc & \#Sent & \#Phrase & \#Token \\ \midrule
Arxiv & 408 & 28.20 & 514.55 & 864.85 \\
ShareGPT & 470 & 27.35 & 389.42 & 689.32 \\
BBC & 294 & 25.63 & 523.96 & 732.54 \\ \bottomrule
\end{tabular}
\vskip -0.1in
\end{table}

\subsection{Tasks and Metrics}

We evaluate Selective Context on four different tasks:

\noindent \textbf{Original Context Reconstruction:} Given a compressed context produced by Selective Context, this task aims to evaluate whether models are able to reconstruct the original context. This task assesses how well the filtered context retains the essential information from the original context. In our experiments, the compressed contexts are used as input, and the original contexts are used as reference answers.

\noindent \textbf{Summarisation:} Given a context, the task is to generate a concise and informative summary that captures the main points of the document. This task aims to evaluate whether the content filtering affects the overall understanding of models on compressed contexts. In our experiments, the input and output are the compressed context and the summaries generated based on the compressed contexts. Summaries based on the \textit{original contexts} are treated as the reference answers.

\noindent \textbf{Question Answering (QA):} Given a document and a set of questions, the task is to generate answers based on the information available in the document. This task aims to evaluate models' fine-grained understanding toward a context. We first generate questions and answers based on the original context, where these answers are treated as reference answers. And then ask LLMs to answer these questions with selective context.

\noindent \textbf{Conversation:} This task is only for the ShareGPT dataset. Given a conversation history and a user query, the task is to generate response to the query based on previous conversation. This task aims to evaluate whether selective context affect the capability of LLMs on conversation. Specifically, we ask LLMs to answer users' last query of ShareGPT conversation instances with selective context applied on previsou conversation history. 

We employ four metrics to assess the performance of our models on the tasks: BLEU, METEOR, ROUGE, and BERTScore. BLEU \cite{papineni2002bleu} calculates n-gram precision, which is the proportion of n-grams in the generated text that are also present in the reference text. METEOR \cite{banerjee2005meteor} take additional features such as synonymy, stemming and word order into consideration, which leads to more comprehensive evaluation. ROUGE \cite{lin2004rouge} focus on how much of the important information in the reference text is present in the generated summary. BERTScore \cite{zhang2019bertscore} is a more recent metric that leverages contextualised embeddings from pre-trained language models like BERT, computing the cosine similarity between the generated text and reference text embeddings to capture semantic similarity more effectively than traditional n-gram-based metrics.

\subsection{Models}

There are two main models were used in our experiments: 

\noindent \textbf{ChatGPT:} We test Selective Context on ChatGPT, which is based on the \texttt{GPT-3.5-turbo} architecture. ChatGPT is a Instruct-tuned language model further improved by RLHF with 175 billion parameters. The base language model of ChatGPT seems to be \texttt{code-davinci-002}\footnote{\url{https://platform.openai.com/docs/model-index-for-researchers}} and more previously \texttt{davinci} which can be found in \cite{brown2020language}. We compare the performance of ChatGPT with and without applying Selective Context to understand its impact on the efficiency and accuracy of the model.

\noindent \textbf{Curie:} \texttt{Curie} is one of the variant of the GPT-3 family with 6.7B of parameters, a smaller version of casual language model \texttt{davinci}. We employ the Curie as the base model $M$ in Selective Context to calculate self-information. Technically, we shall use the same base model of ChatGPT to do content filtering, but our analysis found that the filtering results of \texttt{curie} and \texttt{davinci} are nearly identical, so for the sake of cost, we choose \texttt{curie} instead.

We access the two model via web API provided on the OpenAI platform\footnote{\url{https://platform.openai.com/docs/api-reference}}.

\subsection{Experimental Settings}

We compare different settings to evaluate the effectiveness and analysis trade-offs of Selective Context.

\noindent \textbf{Baseline Comparison:} We compare Selective Context with the original context (without any content reduction) and Random Context, a baseline approach filtering out the same amount of data but does so randomly.

\noindent \textbf{Reduction Ratios:} We experiment with different content reduction ratios in Selective Context: 0.2, 0.35, 0.5, 0.65, and 0.8. These ratios determine the proportion of content to be filtered out, allowing us to study the trade-off between efficiency and performance as the amount of retained information varies.

\noindent \textbf{Lexical Units: } Lexical units are the basic element of content reduction in Selective Context. It can be sentence, phrases, or tokens. But due to the usage limitation of OpenAI web API (\$120 per month), we only test the content filtering on phrase level. It doesn't means self-information based content filtering is not feasible on sentence and token level. We will includes experiments on these two level in the next version.

\input{table1}

% We investigate the impact of content filtering on different lexical units: sentences (sent), phrases, and tokens. By comparing these units, we can better understand the granularity at which Selective Context is most effective.

\noindent \textbf{Self-Information Computation:} Here, we also focus only on sentence-wise self-information calculating, due to the API access limitation as well. Sentence-wise self-information computing means to calculate tokens' self-information sentence by sentence, instead of letting LLMs to process the entire context in one run. Same with lexical units, we will includes experiments testing self-information based on the entire context in the next version.

% We explore two ways of calculating self-information: context-wise computing and sentence-wise computing. Context-wise computing takes the entire context as input and calculates self-information for all tokens in one run. While sentence-wise computing computes self-information sentence by sentence: it take a sentence as input each time and calculate self-information for tokens in it.

We use generation temperature of 0.7 in our experiments.

\section{Results}

\subsection{Comparison to Baselines}

We first compare the performance of Selective Context with different context reduction ratios to the original context on summarisation and QA tasks, as shown in Table \ref{tab:to_original}. The performance drop due the context reduction is shown in parentheses.

As demonstrated in the table, using Selective Context only leads to a marginal drop when the reduction ratio is set to 0.2 or 0.35, despite it significantly reducing the context cost. The BLEU score drops by only 0.05 when 20\% of the content is reduced. And the number is even smaller when it comes to ROUGE-1, where the drop is just 0.03. This indicate a high level of consistency between answers given selective contexts and original contexts when the reduction ratio is 0.2. Selective Context also yields impressive results when 35\% of the content is reduced, with BERT scores around 0.9 and ROUGE-1 scores over 0.5.

The results start to lose control as the reduction ratio rises to 0.5, where the average BLEU score drops 0.17 and the average ROUGE-1 drops 0.12. However, the performance of Selective Context on summarisation and conversation tasks are still acceptable, considering the decrease on BLEU and ROUGE-1 is below 0.1. When 65\% of the context is reduced, the performance of summarisation decreases to 0.114 on BLEU, 0.447 on ROUGE-1, and 0.886 on BERTScore. A reduction ratio of 0.8 tends to be less valuable, as the correctness of answers might not be guaranteed. In summary, the results suggest that Selective Context is very effective in preserving key information during context reduction, and is able to significantly reduce the context cost while preventing big performance loss.

\begin{figure*}[t]
    \centering
    \includegraphics[width=\textwidth]{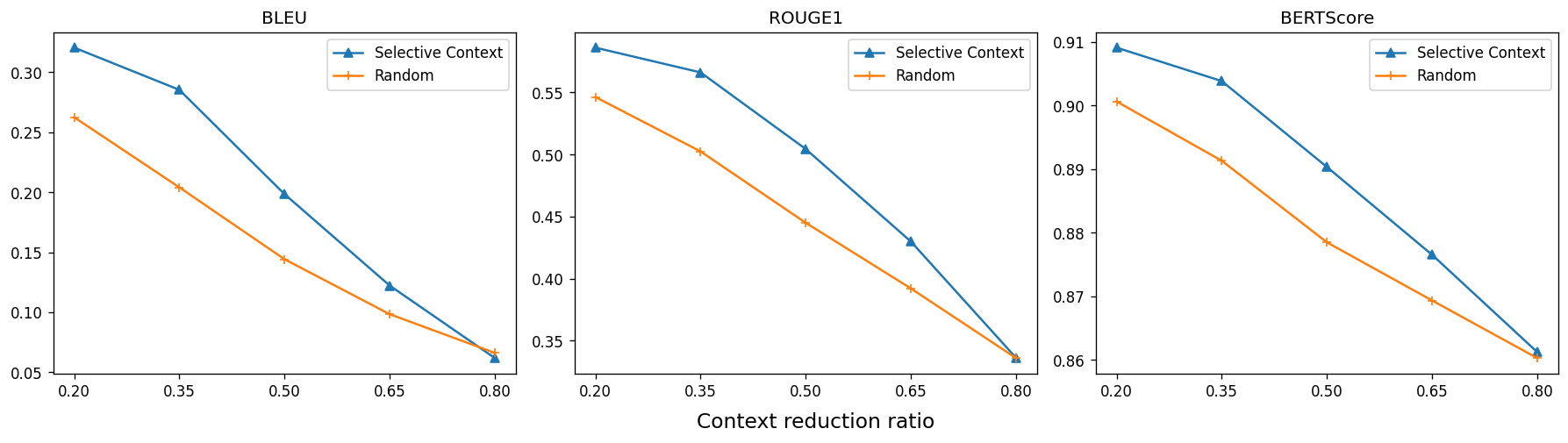}
    \vskip -0.1in
    \caption{Performance of Selective Context compared to the random filtered baselines.}
    \label{fig:random}
\end{figure*}

\begin{figure*}[t]
    \centering
    \includegraphics[width=\textwidth]{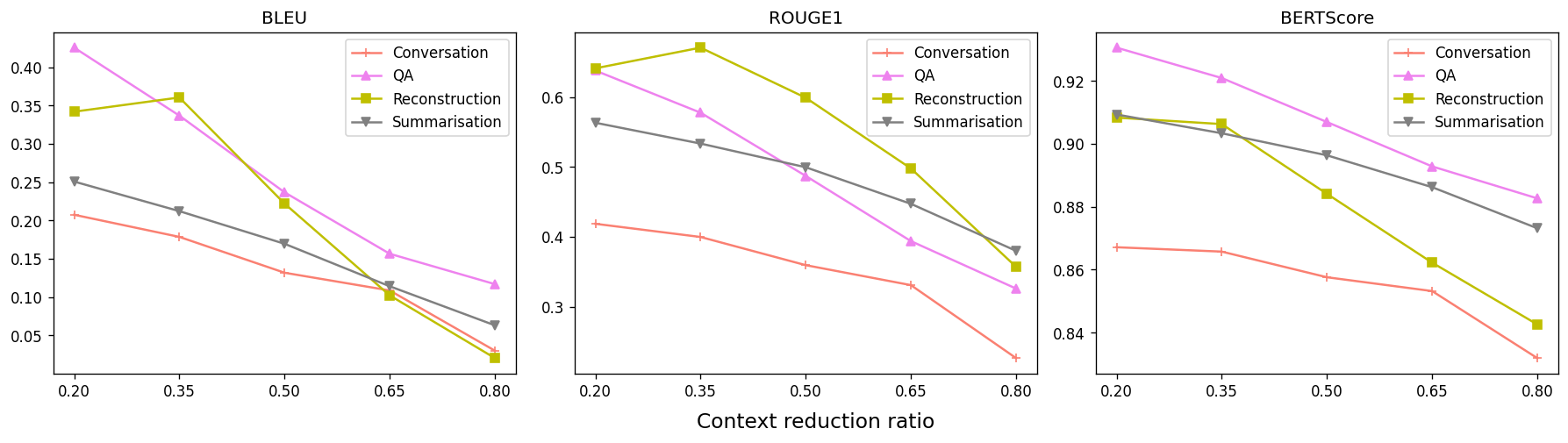}
    \vskip -0.1in
    \caption{Performance of Selective Context on different NLP tasks}
    \label{fig:tasks}
\end{figure*}

We then compare Selective Context to the random filtering baseline, and the results are presented in Figure \ref{fig:random}. Our initial observation reveals that LLMs are quite robust to context reduction. With the random filtering approach, LLMs can achieve over 0.25 BLEU score when 20\% of content is randomly reduced and over 0.5 ROUGE-1 when 35\% of content is randomly reduced. Our proposed method, Selective Context, is even more effective, reaching around 0.3 BLEU score and over 0.55 ROUGE-1 score when the reduction ratio is set to 0.35.

When Selective Context reduces 50\% of content, the performance begins to drop dramatically on BLEU. Nevertheless, the ROUGE-1 and BERT scores remain strong. The rate of performance drop for the random baseline slows between the reduction ratios of 0.5 and 0.65, indicating that the random baseline has already lost a considerable amount of key information after reducing 50\% of content. In contrast, Selective Context does not exhibit this tendency.

When the reduction ratio is set to 0.8, both approaches show similar results, demonstrating that LLMs struggle to handle context with 80\% information loss. Overall, our results show that Selective Context can effectively maximise the utility of fixed context length in LLMs while maintaining strong performance on various tasks.

\begin{figure*}[t]
    \centering
    \includegraphics[width=\textwidth]{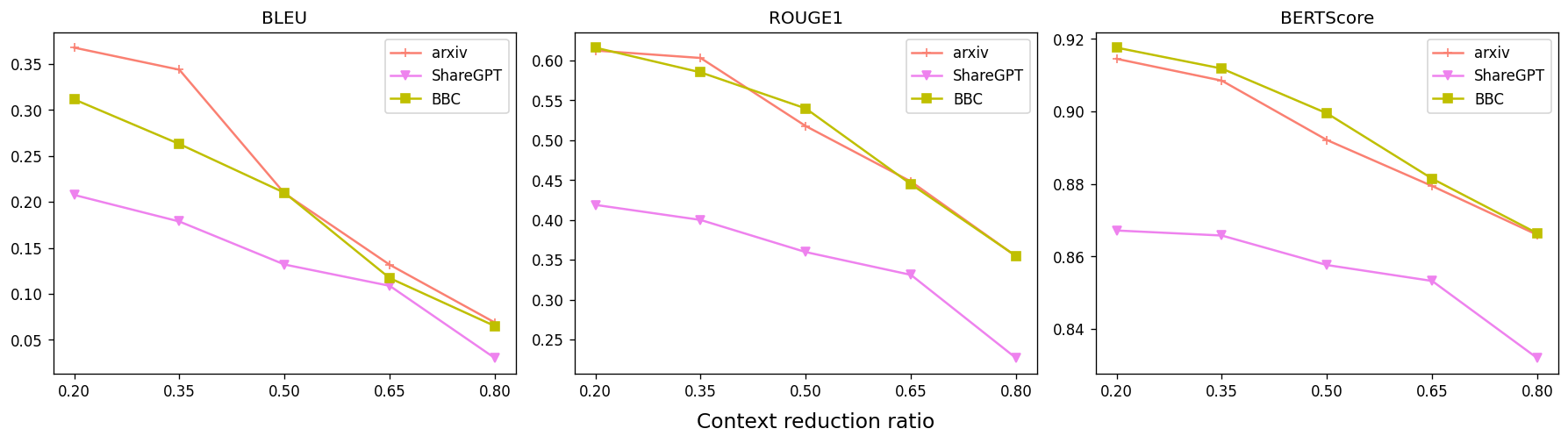}
    \vskip -0.1in
    \caption{Performance of Selective Context on different data sources.}
    \label{fig:datasets}
\end{figure*}

\subsection{Tasks}

In this part, we examine the performances of Selective Context on the three different NLP tasks: summarisation, question answering, and original context reconstruction. The results are as shown in \ref{fig:tasks}.

From the results of the Original Context Reconstruction task (RC), we found that Selective Context allows LLMs to recover most of the key points in the original context when the reduction ratio is set to 0.2 or 0.35, as demonstrated by a rather high ROUGE-1 score of 0.65 and a BERTScore over 0.9. Based on that, it is safe to reduce 35\% of content via Selective Context, which will only leads to minor information loss. However, the performance starts to drop as the reduction ratio increases to 0.5, indicating that partial key information is inevitably lost during the context reduction procedure (ROUGE-1: 0.59, BERTScore: 0.88). The performance decreases dramatically when the reduction ratio is set to 0.8, where we only receive a BLEU score of 0.03 and ROUGE-1 of 0.37.

By comparing the four curves, we found that the summarisation and conversation task seems to be less affected by context reduction. From reduction ratio of 0.2 to 0.8, the BERTScore of summarisation task only show little decrease. On BLEU and ROUGE-1 metrics, the fluctuation of summarisation are also the smallest. The conversation task show the same tendency. On the contrary, reconstruction and QA tasks are significantly influenced by content reduction. This might be because summarisation and conversation tasks focus on overall context understanding, whereas QA and reconstruction tasks require more fine-grained information. As a result, we should be careful when we apply Selective Context for tasks like QA as it might reduce trivial details which is required for some queries. In summary, we observe that Selective Context is quite effective and useful in reducing context cost and can ensure a decent performance when reduction ratio is equal to or below than 0.5.

\subsection{Data Sources}

We also compare how Selective Context perform on different data sources, shown in Figure \ref{fig:datasets}. The performance on ShareGPT is rather lower than the others, but as they are used for different tasks, so we cannot compare their absolute numbers directly. But we are able to recognise that Selective Context works well on arxiv data as long as reduction ratio is equal or lower than 0.35. A considerable performance decrease is found on arxiv as reduction ratio rises to 0.5, which show that the optimal threshold for arxiv data might be between 0.35 and 0.5. For news data, we find the steep performance decrease is between reduction ratio 0.5 to 0.65, for which we seems to be able to use more aggressive context reduction on news data. For conversation tasks, the performance appears stable up to an 80\% context reduction, suggesting that we could potentially have much longer conversations using selective context beyond the fixed context length of LLMs.

\subsection{Case Study}

To have a more straightforward impression of how Selective Context reduce context cost, we present several cases in the Appendix.

\section{Conclusion}

In this paper, we introduced Selective Context to maximise the utility of fixed context length in LLMs. We demonstrated the effectiveness of our method by filtering out less informative content, providing a more compact and efficient context representation for LLMs without sacrificing their performance on various tasks. Our extensive evaluations on arxiv papers, BBC news articles, and conversation transcripts showed that Selective Context can significantly enhance the efficiency of LLMs, enabling them to handle long documents and extended conversations more effectively.


\bibliography{main}
\bibliographystyle{icml2023}

\appendix

% \begin{WideFramedBox}

% \begin{myframedbox}
\begin{figure*}
    \caption{Selective Context with reduction ratio of 0.5 on conversation transcripts.}
    \vskip -0.01in
    \fbox{\parbox[c]{\textwidth}{ 
{\setlength{\fboxsep}{-1pt}
 \input{text_2}
}
   \par }}
% \end{myframedbox}
% \end{WideFramedBox}
\end{figure*}

\begin{figure*}[t]
    \caption{Selective Context with reduction ratio of 0.5 on BBC News.}
    \vskip -0.01in
    \fbox{\parbox[c]{\textwidth}{ 
{\setlength{\fboxsep}{-1pt}
 \input{text_3}
}
   \par }}
% \end{myframedbox}
% \end{WideFramedBox}
\end{figure*}

\end{document}

%% file: text_0.tex
\colorize{23.552335407799475}{INTRODUCTION Continual Learning}
\colorize{15.050009991624608}{ (}
\colorize{0.5782644677890139}{\sdelete{CL}}
\colorize{7.695106268219917}{\sdelete{)}}
\colorize{7.695106268219917}{\sdelete{,}}
\colorize{11.13090181620271}{\sdelete{ also}}
\colorize{1.2520989718578064}{\sdelete{ known}}
\colorize{0.0}{\sdelete{ as}}
\colorize{8.599166557986761}{\sdelete{ Lifelong Learning}}
\colorize{4.203151708838297}{\sdelete{,}}
\colorize{2.374581644022289}{\sdelete{ is}}
\colorize{20.687075258945274}{ a promising learning paradigm}
\colorize{17.045900269063775}{ to}
\colorize{30.036002079520163}{ design}
\colorize{50.310887020929904}{ models}
\colorize{8.964550521989729}{\sdelete{ that}}
\colorize{21.56907410117672}{ have}
\colorize{27.930459470533563}{ to}
\colorize{12.731554061977635}{\sdelete{ learn}}
\colorize{19.962812703073357}{ how}
\colorize{0.0818267481285829}{\sdelete{ to}}
\colorize{13.202391560779375}{\sdelete{ perform}}
\colorize{12.495404958668642}{\sdelete{ multiple tasks}}
\colorize{23.176587979837553}{ across}
\colorize{8.097165367186758}{\sdelete{ different environments}}
\colorize{25.576998609290435}{ over}
\colorize{8.96531610147404}{\sdelete{ their lifetime}}
\colorize{14.05512960198368}{\sdelete{ [}}
\colorize{77.53257825429849}{To}
\colorize{100.0}{ uniform}
\colorize{27.215070848976158}{ the language}
\colorize{17.462422512789907}{ and}
\colorize{38.36594143188037}{ enhance}
\colorize{2.7273554343466366}{\sdelete{ the readability}}
\colorize{2.9080568711890313}{\sdelete{ of}}
\colorize{4.037060820102168}{\sdelete{ the paper}}
\colorize{11.681606351717397}{\sdelete{ we}}
\colorize{20.57882913052647}{ adopt}
\colorize{21.6091754091487}{ the unique term continual learning}
\colorize{5.45516546954231}{\sdelete{ (}}
\colorize{0.027843514679665914}{\sdelete{CL}}
\colorize{5.141316406506657}{\sdelete{)}}
\colorize{5.141316406506657}{\sdelete{.}}
\colorize{1.481087780232099}{\sdelete{]}}
\colorize{1.481087780232099}{\sdelete{.}}
\colorize{45.881450161996696}{Ideal CL models}
\colorize{22.482294449947375}{ in}
\colorize{12.276652829372907}{\sdelete{ the real world}}
\colorize{30.15524576869814}{ should}
\colorize{4.719185851064073}{\sdelete{ be}}
\colorize{43.38486672838565}{ deal}
\colorize{0.08471015710727489}{\sdelete{ with}}
\colorize{62.65258176885338}{ domain shifts}
\colorize{3.252774386202273}{\sdelete{,}}
\colorize{42.88148902114995}{ researchers}
\colorize{5.378334856125341}{\sdelete{ have}}
\colorize{24.4052828965596}{ recently}
\colorize{20.715333420340855}{ started}
\colorize{3.7497948236223384}{\sdelete{ to}}
\colorize{55.1046393186241}{ sample}
\colorize{55.4624828020324}{ tasks}
\colorize{5.112935543833082}{\sdelete{ from}}
\colorize{23.89043034656182}{ two different datasets}
\colorize{8.707566125586402}{\sdelete{.}}
\colorize{29.33293331202289}{For}
\colorize{37.03822014069802}{ instance}
\colorize{0.42517436120092217}{\sdelete{,}}
\colorize{70.34298634973665}{ proposed}
\colorize{35.025427671623405}{ to}
\colorize{50.59324924548084}{ train}
\colorize{20.0646240600965}{ and}
\colorize{30.8930643120124}{ evaluate}
\colorize{14.32279628743447}{\sdelete{ a model}}
\colorize{18.328500429972863}{ on}
\colorize{18.55820223476934}{ Imagenet}
\colorize{46.105279018018074}{ first}
\colorize{11.659700577145538}{\sdelete{ and}}
\colorize{2.7299512715968777}{\sdelete{ then}}
\colorize{59.387039099800006}{ challenge}
\colorize{10.85774508381919}{\sdelete{ its performance}}
\colorize{1.6368666838467467}{\sdelete{ on}}
\colorize{7.456943139535041}{\sdelete{ the}}
\colorize{61.33293296339798}{ Places365}
\colorize{2.4788954639028713}{\sdelete{ dataset}}
\colorize{1.8985536305953028}{\sdelete{.}}
\colorize{46.99659819634475}{considers}
\colorize{46.672769019369376}{ more scenarios}
\colorize{12.085267602731165}{\sdelete{,}}
\colorize{37.161076986645774}{ starting}
\colorize{1.2549350035142675}{\sdelete{ with}}
\colorize{33.11255143396314}{ Imagenet}
\colorize{37.601998772296255}{ or}
\colorize{47.52902699591553}{ Places365}
\colorize{7.518139310785621}{\sdelete{,}}
\colorize{5.723043169101462}{\sdelete{ and}}
\colorize{8.104771260308578}{\sdelete{ then}}
\colorize{11.18163929976799}{\sdelete{ moving}}
\colorize{2.45484091910043}{\sdelete{ on}}
\colorize{0.04733407228536482}{\sdelete{ to}}
\colorize{22.132686237460216}{ the VOC/CUB/Scenes datasets}
\colorize{2.625033888749499}{\sdelete{.}}
\colorize{55.67301137129789}{Few}
\colorize{41.46023557958197}{ works}
\colorize{71.53321154306694}{ propose}
\colorize{32.02627514506558}{ more advanced scenarios}
\colorize{63.21201177136223}{ built}
\colorize{2.9959464952481665}{\sdelete{ on}}
\colorize{15.125679931753902}{ top}
\colorize{0.010878793770251072}{\sdelete{ of}}
\colorize{29.535479934077348}{ more than two datasets}
\colorize{4.123419854370624}{\sdelete{.}}

%% file: text_1.tex
\colorize{23.552335407799475}{INTRODUCTION Continual Learning}
\colorize{15.050009991624608}{ (}
\colorize{20.687075258945274}{ a promising learning paradigm}
\colorize{17.045900269063775}{ to}
\colorize{30.036002079520163}{ design}
\colorize{50.310887020929904}{ models}
\colorize{21.56907410117672}{ have}
\colorize{27.930459470533563}{ to}
\colorize{19.962812703073357}{ how}
\colorize{23.176587979837553}{ across}
\colorize{25.576998609290435}{ over}
\colorize{77.53257825429849}{To}
\colorize{100.0}{ uniform}
\colorize{27.215070848976158}{ the language}
\colorize{17.462422512789907}{ and}
\colorize{38.36594143188037}{ enhance}
\colorize{20.57882913052647}{ adopt}
\colorize{21.6091754091487}{ the unique term continual learning}
\colorize{45.881450161996696}{Ideal CL models}
\colorize{22.482294449947375}{ in}
\colorize{30.15524576869814}{ should}
\colorize{43.38486672838565}{ deal}
\colorize{62.65258176885338}{ domain shifts}
\colorize{42.88148902114995}{ researchers}
\colorize{24.4052828965596}{ recently}
\colorize{20.715333420340855}{ started}
\colorize{55.1046393186241}{ sample}
\colorize{55.4624828020324}{ tasks}
\colorize{23.89043034656182}{ two different datasets}
\colorize{29.33293331202289}{For}
\colorize{37.03822014069802}{ instance}
\colorize{70.34298634973665}{ proposed}
\colorize{35.025427671623405}{ to}
\colorize{50.59324924548084}{ train}
\colorize{20.0646240600965}{ and}
\colorize{30.8930643120124}{ evaluate}
\colorize{18.328500429972863}{ on}
\colorize{18.55820223476934}{ Imagenet}
\colorize{46.105279018018074}{ first}
\colorize{59.387039099800006}{ challenge}
\colorize{61.33293296339798}{ Places365}
\colorize{46.99659819634475}{considers}
\colorize{46.672769019369376}{ more scenarios}
\colorize{37.161076986645774}{ starting}
\colorize{33.11255143396314}{ Imagenet}
\colorize{37.601998772296255}{ or}
\colorize{47.52902699591553}{ Places365}
\colorize{22.132686237460216}{ the VOC/CUB/Scenes datasets}
\colorize{55.67301137129789}{Few}
\colorize{41.46023557958197}{ works}
\colorize{71.53321154306694}{ propose}
\colorize{32.02627514506558}{ more advanced scenarios}
\colorize{63.21201177136223}{ built}
\colorize{15.125679931753902}{ top}
\colorize{29.535479934077348}{ more than two datasets}

%% file: table1.tex
\begin{table*}[t]
    \centering
\caption{Comparing Selective Context with different context reduction ratio to the Original context, on Summarisation and QA task. The performance drop are shown in parentheses.}
    \label{tab:to_original}
\vskip 0.1in
\resizebox{\textwidth}{!}{
\begin{tabular}{llrrrrrrrr}
\toprule
& & & & \multicolumn{3}{c}{ROUGE} & \multicolumn{3}{c}{BERTScore} \\
\cmidrule(lr){5-7}
\cmidrule(lr){8-10}
Method & Task & BLEU & METEOR & rouge1 & rouge2 & rougeL & Precision & Recall & F1 \\
\midrule
\multirow[t]{4}{*}{Original} & Summarisation & .274 & .481 & .570 & .321 & .416 & .912 & .911 & .911 \\
 & QA & .529 & .664 & .690 & .581 & .664 & .941 & .939 & .940 \\
 & Conversation & .238 & .343 & .451 & .249 & .332 & .878 & .878 & .877 \\
 & Avg. & .347 & .496 & .571 & .383 & .471 & .910 & .909 & .909 \\
\cmidrule(lr){1-2}
\cmidrule(lr){3-10}
\multirow[t]{4}{*}{SC-0.2} & Summarisation & .251 (.02) & .475 (.01) & .563 (.01) & .305 (.02) & .402 (.01) & .910 (.002) & .909 (.002) & .909 (.002) \\
 & QA & .426 (.10) & .601 (.06) & .638 (.05) & .502 (.08) & .605 (.06) & .933 (.008) & .929 (.010) & .931 (.009) \\
 & Conversation & .208 (.03) & .305 (.04) & .419 (.03) & .230 (.02) & .307 (.02) & .873 (.005) & .862 (.015) & .867 (.010) \\
 & Avg. & .295 (.05) & .460 (.04) & .540 (.03) & .346 (.04) & .438 (.03) & .905 (.005) & .900 (.009) & .902 (.007) \\
\cmidrule(lr){1-2}
\cmidrule(lr){3-10}
\multirow[t]{4}{*}{SC-0.35} & Summarisation & .212 (.06) & .442 (.04) & .533 (.04) & .265 (.06) & .363 (.05) & .905 (.007) & .902 (.009) & .903 (.008) \\
 & QA & .337 (.19) & .531 (.13) & .578 (.11) & .420 (.16) & .539 (.13) & .925 (.017) & .918 (.021) & .921 (.019) \\
 & Conversation & .179 (.06) & .290 (.05) & .400 (.05) & .198 (.05) & .285 (.05) & .871 (.007) & .861 (.016) & .866 (.012) \\
 & Avg. & .243 (.10) & .421 (.08) & .504 (.07) & .294 (.09) & .396 (.07) & .900 (.010) & .894 (.015) & .897 (.013) \\
\cmidrule(lr){1-2}
\cmidrule(lr){3-10}
\multirow[t]{4}{*}{SC-0.5} & Summarisation & .170 (.10) & .397 (.08) & .500 (.07) & .226 (.10) & .331 (.09) & .900 (.012) & .893 (.018) & .896 (.015) \\
 & QA & .237 (.29) & .434 (.23) & .487 (.20) & .321 (.26) & .447 (.22) & .912 (.029) & .903 (.036) & .907 (.033) \\
 & Conversation & .132 (.11) & .254 (.09) & .360 (.09) & .163 (.09) & .254 (.08) & .867 (.012) & .850 (.028) & .858 (.020) \\
 & Avg. & .179 (.17) & .362 (.13) & .449 (.12) & .237 (.15) & .344 (.13) & .893 (.018) & .882 (.027) & .887 (.023) \\
\cmidrule(lr){1-2}
\cmidrule(lr){3-10}
\multirow[t]{4}{*}{SC-0.65} & Summarisation & .114 (.16) & .335 (.15) & .447 (.12) & .168 (.15) & .281 (.13) & .893 (.019) & .880 (.031) & .886 (.025) \\
 & QA & .157 (.37) & .336 (.33) & .394 (.30) & .227 (.35) & .353 (.31) & .899 (.042) & .888 (.051) & .893 (.047) \\
 & Conversation & .109 (.13) & .227 (.12) & .331 (.12) & .139 (.11) & .225 (.11) & .864 (.014) & .843 (.034) & .853 (.024) \\
 & Avg. & .127 (.22) & .299 (.20) & .391 (.18) & .178 (.21) & .287 (.18) & .885 (.025) & .870 (.039) & .877 (.032) \\
\cmidrule(lr){1-2}
\cmidrule(lr){3-10}
\multirow[t]{4}{*}{SC-0.8} & Summarisation & .063 (.21) & .259 (.22) & .380 (.19) & .114 (.21) & .231 (.19) & .884 (.028) & .863 (.048) & .873 (.038) \\
 & QA & .117 (.41) & .272 (.39) & .326 (.36) & .172 (.41) & .289 (.37) & .890 (.051) & .876 (.063) & .883 (.057) \\
 & Conversation & .030 (.21) & .142 (.20) & .227 (.22) & .081 (.17) & .154 (.18) & .849 (.029) & .816 (.061) & .832 (.046) \\
 & Avg. & .070 (.28) & .224 (.27) & .311 (.26) & .122 (.26) & .225 (.25) & .874 (.036) & .852 (.057) & .863 (.047) \\
\bottomrule
\end{tabular}
}

\end{table*}

%% file: text_2.tex
\textbf{human:}
\colorize{33.00506938752763}{Poly Ether Ether Ketone}

\textbf{gpt:}
\colorize{33.00505750877941}{Poly Ether Ether Ketone}
\colorize{1.9796403375875762}{\sdelete{ (PEEK}}
\colorize{0.6815342603230211}{\sdelete{)}}
\colorize{9.45713793616194}{\sdelete{ is}}
\colorize{5.546771057419689}{\sdelete{ a high-performance engineering thermoplastic}}
\colorize{12.81793588844209}{ that}
\colorize{7.890492916024881}{\sdelete{ is}}
\colorize{35.46091188457514}{ known}
\colorize{1.2589998692545694}{\sdelete{ for}}
\colorize{7.335625698630547}{\sdelete{ its excellent mechanical, thermal, and chemical properties}}
\colorize{1.1174835105519862}{\sdelete{.}}
\colorize{43.47965339780286}{PEEK}
\colorize{27.333316710272022}{ is}
\colorize{8.998162052695486}{\sdelete{ a semi-crystalline polymer}}
\colorize{9.700613052336264}{\sdelete{ that}}
\colorize{6.829529569411323}{\sdelete{ is}}
\colorize{32.7948650590358}{ characterized}
\colorize{0.31035616960285106}{\sdelete{ by}}
\colorize{10.98431039893642}{ its high strength}
\colorize{6.985325453526031}{\sdelete{,}}
\colorize{15.945114101082558}{ stiffness}
\colorize{2.8429063666033723}{\sdelete{,}}
\colorize{4.011347917351205}{\sdelete{ and}}
\colorize{16.107397848675948}{ toughness}
\colorize{13.633649777559546}{,}
\colorize{11.550317187096748}{ as}
\colorize{0.23837009002505827}{\sdelete{ well}}
\colorize{0.013425727683452626}{\sdelete{ as}}
\colorize{9.89125297401985}{\sdelete{ its excellent resistance}}
\colorize{0.10294800854143307}{\sdelete{ to}}
\colorize{17.36811031553621}{ heat}
\colorize{4.579880433689891}{\sdelete{,}}
\colorize{8.098140656745052}{\sdelete{ chemicals}}
\colorize{0.6200040504149945}{\sdelete{,}}
\colorize{2.4548175766121343}{\sdelete{ and}}
\colorize{5.458416425920937}{\sdelete{ abrasion}}
\colorize{0.38117930229803765}{\sdelete{.}}
\colorize{43.47965339780286}{PEEK}
\colorize{27.333316710272022}{ is}
\colorize{38.32002012001784}{ widely}
\colorize{1.5642067480595565}{\sdelete{ used}}
\colorize{4.16281820635378}{\sdelete{ in}}
\colorize{14.734043968950223}{ various industries}
\colorize{10.42352577744524}{\sdelete{,}}
\colorize{6.141516581661956}{\sdelete{ including}}
\colorize{12.18496732652558}{ aerospace}
\colorize{1.3553484707548011}{\sdelete{,}}
\colorize{7.085270053438399}{\sdelete{ automotive}}
\colorize{0.6131492780540048}{\sdelete{,}}
\colorize{13.504288466259812}{ medical}
\colorize{2.285162779364948}{\sdelete{,}}
\colorize{9.706941536383937}{\sdelete{ and}}
\colorize{27.09437705954652}{ oil}
\colorize{4.758462346801277}{\sdelete{ and}}
\colorize{0.08996200285592026}{\sdelete{ gas}}
\colorize{3.297071067099297}{\sdelete{.}}
\colorize{29.241671230491935}{It}
\colorize{11.375840913031187}{ is}
\colorize{42.74530893397442}{ used}
\colorize{10.954674540831418}{ in}
\colorize{42.42335625757762}{ applications}
\colorize{12.56314483266614}{ that}
\colorize{5.286199020696881}{\sdelete{ require}}
\colorize{21.42653522829422}{ high-performance materials}
\colorize{10.944794883836579}{,}
\colorize{8.472235310117817}{\sdelete{ such}}
\colorize{0.026946503656375777}{\sdelete{ as}}
\colorize{24.144411030774616}{ structural components}
\colorize{5.678107811909931}{\sdelete{,}}
\colorize{23.978462490131875}{ bearings}
\colorize{1.568250067257551}{\sdelete{,}}
\colorize{27.96473879364702}{ seals}
\colorize{2.0118205803954776}{\sdelete{,}}
\colorize{7.129747498480193}{\sdelete{ and}}
\colorize{16.220897793710794}{ electrical insulation}
\colorize{2.5055534036549942}{\sdelete{.}}
\colorize{34.4140380532083}{One}
\colorize{3.3528108479575214}{\sdelete{ of}}
\colorize{25.45547305783537}{ the unique properties}
\colorize{0.4235287206479136}{\sdelete{ of}}
\colorize{40.13705147142152}{ PEEK}
\colorize{3.347059740801713}{\sdelete{ is}}
\colorize{5.6899211896575554}{\sdelete{ its ability}}
\colorize{0.05752638978912797}{\sdelete{ to}}
\colorize{10.386004995536618}{\sdelete{ withstand}}
\colorize{6.4495834807760835}{\sdelete{ high temperatures}}
\colorize{5.8771683632086}{\sdelete{.}}
\colorize{29.241671230491935}{It}
\colorize{20.1400392473689}{ has}
\colorize{33.17796838571597}{ a melting point}
\colorize{1.4585120993704146}{\sdelete{ of}}
\colorize{20.705343085077853}{ around 343°C}
\colorize{11.12670595932834}{ (}
\colorize{19.657163774301313}{649°}
\colorize{0.07393387963558117}{\sdelete{F}}
\colorize{7.000988218078059}{\sdelete{)}}
\colorize{4.968525682022675}{\sdelete{ and}}
\colorize{21.364181605723466}{ can}
\colorize{46.64271750803231}{ operate}
\colorize{4.216102236565074}{\sdelete{ at}}
\colorize{8.048457232007573}{\sdelete{ temperatures}}
\colorize{8.54365099187961}{\sdelete{ up}}
\colorize{0.027690080920836482}{\sdelete{ to}}
\colorize{33.52464230604439}{ 250}
\colorize{0.41925401489004116}{\sdelete{°}}
\colorize{0.07161841401673356}{\sdelete{C}}
\colorize{0.663956576810458}{\sdelete{ (}}
\colorize{1.4224263549167746}{\sdelete{482}}
\colorize{0.07520992365661339}{\sdelete{°}}
\colorize{0.049028272301979925}{\sdelete{F}}
\colorize{11.591498715311353}{)}
\colorize{16.01623405929072}{ for}
\colorize{7.935481869428071}{\sdelete{ long periods}}
\colorize{3.0244156938653304}{\sdelete{ of}}
\colorize{0.05063982870256275}{\sdelete{ time}}
\colorize{13.049301903315158}{ without}
\colorize{15.600390088480385}{ losing}
\colorize{12.42923691180893}{ its mechanical properties}
\colorize{1.5932746301688196}{\sdelete{.}}
\colorize{43.47965339780286}{PEEK}
\colorize{27.333316710272022}{ is}
\colorize{32.69362736897419}{ also}
\colorize{35.62908457762913}{ highly}
\colorize{10.3760518745853}{\sdelete{ resistant}}
\colorize{0.26661852528009333}{\sdelete{ to}}
\colorize{27.442602439023673}{ chemicals}
\colorize{7.917467262147218}{\sdelete{ and}}
\colorize{23.48015585074795}{ can}
\colorize{9.51241582795819}{\sdelete{ withstand}}
\colorize{21.81253150954877}{ exposure}
\colorize{0.29103006260332637}{\sdelete{ to}}
\colorize{15.372578670721943}{ harsh environments}
\colorize{21.162720401745226}{ without}
\colorize{15.184149536820232}{ degrading}
\colorize{2.6729564683784597}{\sdelete{.}}
\colorize{29.241671230491935}{It}
\colorize{11.375840913031187}{ is}
\colorize{84.22533365527806}{ resistant}
\colorize{0.5033357294523197}{\sdelete{ to}}
\colorize{10.877464080174134}{ most organic solvents}
\colorize{6.888513032978527}{\sdelete{,}}
\colorize{19.188477428624832}{ acids}
\colorize{3.309592500420696}{\sdelete{,}}
\colorize{9.424580329347076}{\sdelete{ and}}
\colorize{5.467958561214005}{\sdelete{ bases}}
\colorize{11.158760524814085}{,}
\colorize{5.859023388533791}{\sdelete{ and}}
\colorize{14.754367336707453}{ can}
\colorize{3.643047889189826}{\sdelete{ be}}
\colorize{12.894957243628893}{ used}
\colorize{5.2680625130543435}{\sdelete{ in}}
\colorize{21.830287374284595}{ applications}
\colorize{6.533977224154876}{\sdelete{ where}}
\colorize{17.590960114640648}{ other plastics}
\colorize{17.34751052407832}{ would}
\colorize{23.75128641108181}{ fail}
\colorize{1.5733374462438336}{\sdelete{.}}
\colorize{51.61521878880823}{Overall}
\colorize{10.874567699472426}{\sdelete{,}}
\colorize{49.129741698978975}{ PEEK}
\colorize{7.848667021751317}{\sdelete{ is}}
\colorize{12.45714437469375}{ a versatile material}
\colorize{6.666160672020088}{\sdelete{ that}}
\colorize{14.177444925250729}{ offers}
\colorize{7.857928110676148}{\sdelete{ excellent mechanical, thermal, and chemical properties}}
\colorize{13.083787627699687}{,}
\colorize{9.634307473581192}{\sdelete{ making}}
\colorize{0.21333167479996173}{\sdelete{ it}}
\colorize{9.010228050398837}{\sdelete{ an}}
\colorize{7.45451297280393}{\sdelete{ ideal}}
\colorize{7.040247854482024}{\sdelete{ choice}}
\colorize{0.5458940220174435}{\sdelete{ for}}
\colorize{4.216195726713081}{\sdelete{ a wide range}}
\colorize{0.006725578896153351}{\sdelete{ of}}
\colorize{8.608280440661506}{\sdelete{ high-performance applications}}
\colorize{1.5471918109051526}{\sdelete{.}}

\textbf{human:}
\colorize{60.28072203306964}{Application}
\colorize{40.599376336527506}{ in}
\colorize{35.47748662943777}{ Membrane manufacturing}

\textbf{gpt:}
\colorize{43.47968172500642}{PEEK}
\colorize{27.33330488132988}{ is}
\colorize{32.693636707612725}{ also}
\colorize{13.748320787866767}{ used}
\colorize{6.113334438289006}{\sdelete{ in}}
\colorize{45.671750309699064}{ membrane manufacturing}
\colorize{36.46993123334968}{ due}
\colorize{0.061096650913459616}{\sdelete{ to}}
\colorize{10.705554010136979}{\sdelete{ its unique properties}}
\colorize{5.7867441939954}{\sdelete{.}}
\colorize{32.15729856655493}{Membranes}
\colorize{30.13575061616943}{ made}
\colorize{5.250680193859947}{\sdelete{ from}}
\colorize{26.46554769351697}{ PEEK}
\colorize{27.843032053067024}{ offer}
\colorize{11.212900638396208}{ excellent chemical resistance}
\colorize{7.265650837362116}{\sdelete{ and}}
\colorize{21.613404342628545}{ durability}
\colorize{9.901226307935465}{\sdelete{,}}
\colorize{14.366888546585365}{ making}
\colorize{1.7863102654298155}{\sdelete{ them}}
\colorize{15.157414882420309}{ suitable}
\colorize{0.3643286193439586}{\sdelete{ for}}
\colorize{12.685890029851299}{ use}
\colorize{1.3762301023236945}{\sdelete{ in}}
\colorize{11.184278354612806}{ a variety}
\colorize{0.004023318698100379}{\sdelete{ of}}
\colorize{5.523823853374148}{\sdelete{ applications}}
\colorize{12.433989593990093}{,}
\colorize{2.6216523540519083}{\sdelete{ including}}
\colorize{20.473534169707833}{ water treatment}
\colorize{4.13886397593388}{\sdelete{,}}
\colorize{17.91814896566439}{ gas separation}
\colorize{1.6654073325973815}{\sdelete{,}}
\colorize{8.717911232107047}{\sdelete{ and}}
\colorize{18.86209932706996}{ biomedical applications}
\colorize{0.8018364696672504}{\sdelete{.}}
\colorize{27.3264783365603}{In}
\colorize{46.64397666779486}{ water treatment}
\colorize{4.441614170228083}{\sdelete{,}}
\colorize{43.313239066622494}{ PEEK membranes}
\colorize{3.927901267563907}{\sdelete{ are}}
\colorize{7.695965962447329}{\sdelete{ used}}
\colorize{6.825465393920599}{\sdelete{ to}}
\colorize{19.51205627327863}{ filter}
\colorize{10.094574102753914}{\sdelete{ out}}
\colorize{10.734109053647671}{\sdelete{ impurities}}
\colorize{14.163412064411126}{,}
\colorize{18.821379905818894}{ including}
\colorize{10.572869907483604}{\sdelete{ bacteria}}
\colorize{7.440036837921833}{\sdelete{ and}}
\colorize{5.783828422014202}{\sdelete{ viruses}}
\colorize{5.533434868867111}{\sdelete{,}}
\colorize{6.3258034067971955}{\sdelete{ from}}
\colorize{4.927284386521443}{\sdelete{ water}}
\colorize{2.7768676861376487}{\sdelete{.}}
\colorize{59.00337838484121}{These membranes}
\colorize{5.352871670203998}{\sdelete{ are}}
\colorize{33.09593902997043}{ highly}
\colorize{16.791228994089042}{ effective}
\colorize{8.68942340396603}{\sdelete{ at}}
\colorize{9.646001316755559}{\sdelete{ removing}}
\colorize{11.413680178000702}{ impurities}
\colorize{8.551095131944345}{\sdelete{ and}}
\colorize{30.460297566351436}{ can}
\colorize{3.674076449588448}{\sdelete{ be}}
\colorize{2.381813703723375}{\sdelete{ used}}
\colorize{8.178875054613641}{\sdelete{ in}}
\colorize{7.955096372262853}{\sdelete{ a variety}}
\colorize{0.002235849118269318}{\sdelete{ of}}
\colorize{13.121956967678344}{ water treatment applications}
\colorize{12.734888621096166}{,}
\colorize{3.5018449351965}{\sdelete{ including}}
\colorize{9.28615040313205}{\sdelete{ desalination}}
\colorize{3.8649541122566218}{\sdelete{,}}
\colorize{11.573147294467315}{ wastewater treatment}
\colorize{2.0818235129195948}{\sdelete{,}}
\colorize{7.716628634071357}{\sdelete{ and}}
\colorize{19.777708165671566}{ drinking}
\colorize{3.4866462345715985}{\sdelete{ water purification}}
\colorize{0.5971109757176617}{\sdelete{.}}
\colorize{27.3264783365603}{In}
\colorize{58.80620392723878}{ gas separation}
\colorize{5.035168074340935}{\sdelete{,}}
\colorize{36.83133467695247}{ PEEK membranes}
\colorize{3.8787977722794915}{\sdelete{ are}}
\colorize{6.937695284114594}{\sdelete{ used}}
\colorize{7.681365934960661}{\sdelete{ to}}
\colorize{4.311355727053023}{\sdelete{ separate}}
\colorize{10.529577224384035}{\sdelete{ gases}}
\colorize{25.447843099616303}{ based}
\colorize{0.2039243474649807}{\sdelete{ on}}
\colorize{11.354685368817776}{ their molecular size}
\colorize{12.977733690457283}{ and}
\colorize{25.35559851846455}{ shape}
\colorize{1.9283642806816919}{\sdelete{.}}
\colorize{59.00337838484121}{These membranes}
\colorize{5.352871670203998}{\sdelete{ are}}
\colorize{33.09593902997043}{ highly}
\colorize{16.791228994089042}{ effective}
\colorize{8.68942340396603}{\sdelete{ at}}
\colorize{22.590798703289988}{ separating}
\colorize{25.817183140822607}{ gases}
\colorize{8.866536286978164}{\sdelete{ and}}
\colorize{26.066073422195817}{ can}
\colorize{2.778315486398629}{\sdelete{ be}}
\colorize{2.928234565282838}{\sdelete{ used}}
\colorize{6.608341425390331}{\sdelete{ in}}
\colorize{6.508229663848175}{\sdelete{ a variety}}
\colorize{0.004656426094443118}{\sdelete{ of}}
\colorize{2.91072866477182}{\sdelete{ applications}}
\colorize{11.60097992372867}{,}
\colorize{4.441299769397382}{\sdelete{ including}}
\colorize{13.51820490540567}{ air separation}
\colorize{4.105149249546857}{\sdelete{,}}
\colorize{14.59020264204527}{ natural gas processing}
\colorize{1.5249799214056476}{\sdelete{,}}
\colorize{8.645011329393986}{\sdelete{ and}}
\colorize{13.57014333119036}{ hydrogen purification}
\colorize{0.843816389219065}{\sdelete{.}}
\colorize{27.3264783365603}{In}
\colorize{48.34841236566191}{ biomedical applications}
\colorize{2.17498440128447}{\sdelete{,}}
\colorize{39.97835147824752}{ PEEK membranes}
\colorize{4.069124268501289}{\sdelete{ are}}
\colorize{9.133786331712503}{\sdelete{ used}}
\colorize{8.548678292291314}{\sdelete{ as}}
\colorize{18.518231109378448}{ implantable devices}
\colorize{10.188368275326686}{\sdelete{,}}
\colorize{22.685612655190855}{ including}
\colorize{19.468192688076584}{ cardiovascular implants}
\colorize{4.540540671767292}{\sdelete{,}}
\colorize{5.36468811569438}{\sdelete{ orthopedic implants}}
\colorize{1.1432321292377208}{\sdelete{,}}
\colorize{9.303850410357912}{\sdelete{ and}}
\colorize{10.320796042728524}{\sdelete{ drug delivery systems}}
\colorize{1.2872924575926012}{\sdelete{.}}
\colorize{43.47965339780286}{PEEK}
\colorize{27.333316710272022}{ is}
\colorize{11.066880411528285}{ a biocompatible material}
\colorize{7.1280709015750885}{\sdelete{ that}}
\colorize{7.559380280376401}{\sdelete{ is}}
\colorize{31.95409747045484}{ well}
\colorize{8.905452261484015}{\sdelete{-}}
\colorize{7.516340291081917}{\sdelete{tolerated}}
\colorize{3.235841785479443}{\sdelete{ by}}
\colorize{4.890752827915628}{\sdelete{ the human body}}
\colorize{10.498120331764577}{\sdelete{ and}}
\colorize{19.702949251314006}{ does}
\colorize{0.014703193667845876}{\sdelete{ not}}
\colorize{11.242856832555514}{ cause}
\colorize{14.041601353655436}{ adverse reactions}
\colorize{18.814519119374687}{ or}
\colorize{38.624394365810986}{ rejection}
\colorize{4.556716625635651}{\sdelete{.}}
\colorize{51.61521878880823}{Overall}
\colorize{10.874567699472426}{\sdelete{,}}
\colorize{49.129741698978975}{ PEEK}
\colorize{7.848667021751317}{\sdelete{ is}}
\colorize{12.45714437469375}{ a versatile material}
\colorize{6.666160672020088}{\sdelete{ that}}
\colorize{9.98726878794773}{\sdelete{ is}}
\colorize{21.349225464820638}{ well}
\colorize{10.806542566074985}{\sdelete{-}}
\colorize{0.11846501624577274}{\sdelete{suited}}
\colorize{2.775099446059714}{\sdelete{ for}}
\colorize{17.81373769494041}{ use}
\colorize{1.4124194434074901}{\sdelete{ in}}
\colorize{50.57295917874186}{ membrane manufacturing}
\colorize{23.849630990925068}{ due}
\colorize{0.08106068031135605}{\sdelete{ to}}
\colorize{9.976013113693455}{\sdelete{ its excellent chemical resistance}}
\colorize{4.996504243075547}{\sdelete{,}}
\colorize{29.12595616039643}{ durability}
\colorize{1.7454648659449044}{\sdelete{,}}
\colorize{3.0818584065893293}{\sdelete{ and}}
\colorize{6.039165702540301}{\sdelete{ biocompatibility}}
\colorize{1.1059539651582562}{\sdelete{.}}

\textbf{human:}
\colorize{33.27215602625905}{Properties}
\colorize{10.162371995946987}{\sdelete{ of}}
\colorize{38.99173531579475}{ PEEK}
\colorize{70.52169717861204}{ required}
\colorize{3.3614972759778614}{\sdelete{ for}}
\colorize{29.03410508768241}{ Membrane Manufacturing}

\textbf{gpt:}
\colorize{31.625919035034265}{There}
\colorize{6.28068719887956}{\sdelete{ are}}
\colorize{33.5774485715082}{ several properties}
\colorize{8.606178780059052}{\sdelete{ of}}
\colorize{44.710981838510634}{ PEEK}
\colorize{7.6566390878459085}{\sdelete{ that}}
\colorize{6.173520967529966}{\sdelete{ make}}
\colorize{0.6123619063102671}{\sdelete{ it}}
\colorize{12.879670514921903}{ an}
\colorize{7.601912175723265}{\sdelete{ ideal}}
\colorize{4.646395571493705}{\sdelete{ material}}
\colorize{0.9900260363617254}{\sdelete{ for}}
\colorize{45.369673057697746}{ membrane manufacturing}
\colorize{16.83431249168541}{,}
\colorize{8.472784422063715}{\sdelete{ including}}
\colorize{11.797144251929478}{:}
\colorize{27.771308663165485}{ Chemical resistance}
\colorize{14.375628267102947}{:}
\colorize{1.5164029939027455}{\sdelete{ PEEK}}
\colorize{12.426994331150972}{ has}
\colorize{7.012399878724675}{\sdelete{ excellent resistance}}
\colorize{0.23516439689764118}{\sdelete{ to}}
\colorize{7.635221397672538}{\sdelete{ a wide range}}
\colorize{0.016378496860224988}{\sdelete{ of}}
\colorize{4.082478148061758}{\sdelete{ chemicals}}
\colorize{6.19531361441681}{\sdelete{,}}
\colorize{3.701283016522373}{\sdelete{ including}}
\colorize{10.106173936967021}{\sdelete{ acids}}
\colorize{1.0031066673586406}{\sdelete{,}}
\colorize{5.431059110055957}{\sdelete{ bases}}
\colorize{0.6315519991523992}{\sdelete{,}}
\colorize{11.394680059837148}{ and}
\colorize{4.346624274601279}{\sdelete{ organic solvents}}
\colorize{1.9360425092857216}{\sdelete{.}}
\colorize{33.9268898417227}{This property}
\colorize{42.59907457056843}{ makes}
\colorize{15.889833718982699}{ it}
\colorize{37.26212794033071}{ ideal}
\colorize{0.8187034205296586}{\sdelete{ for}}
\colorize{27.429726324210684}{ use}
\colorize{14.093595157566641}{ in}
\colorize{46.06579509657559}{ applications}
\colorize{5.553168470472019}{\sdelete{ where}}
\colorize{34.11194741582508}{ the membrane}
\colorize{25.650857007895073}{ may}
\colorize{25.838988861803397}{ come}
\colorize{3.772691227373337}{\sdelete{ into}}
\colorize{1.4332664597562117}{\sdelete{ contact}}
\colorize{0.07994252776492594}{\sdelete{ with}}
\colorize{16.006375414232586}{ harsh chemicals}
\colorize{10.463138414385853}{\sdelete{.}}
\colorize{36.76540036096051}{Thermal stability}
\colorize{25.608016314900635}{:}
\colorize{37.83937595898438}{ PEEK}
\colorize{9.934958092902214}{\sdelete{ is}}
\colorize{21.60990795636079}{ highly}
\colorize{9.496328466640739}{\sdelete{ stable}}
\colorize{14.653491363245127}{ at}
\colorize{4.606929924723645}{\sdelete{ high temperatures}}
\colorize{10.036884974487753}{\sdelete{,}}
\colorize{25.590477106578835}{ making}
\colorize{0.691028480576697}{\sdelete{ it}}
\colorize{10.373470052318137}{\sdelete{ ideal}}
\colorize{0.3262118579755085}{\sdelete{ for}}
\colorize{7.714427205680548}{\sdelete{ use}}
\colorize{1.1517970926995071}{\sdelete{ in}}
\colorize{13.717639624821668}{ applications}
\colorize{8.50609347799275}{\sdelete{ where}}
\colorize{37.66208439668826}{ the membrane}
\colorize{19.650635443388975}{ may}
\colorize{1.7636742772256107}{\sdelete{ be}}
\colorize{1.7459155486406364}{\sdelete{ exposed}}
\colorize{0.13038479781829693}{\sdelete{ to}}
\colorize{5.110867931240076}{\sdelete{ elevated temperatures}}
\colorize{22.761141695939667}{ during}
\colorize{11.324733034629064}{ use}
\colorize{11.958988460935787}{,}
\colorize{43.07669301271634}{ cleaning}
\colorize{3.5124175191706852}{\sdelete{,}}
\colorize{3.1846166137773144}{\sdelete{ or}}
\colorize{7.641292364612083}{\sdelete{ sterilization}}
\colorize{0.9992933899565747}{\sdelete{.}}
\colorize{32.3401038685678}{Mechanical}
\colorize{49.71741008179869}{ strength}
\colorize{19.870256563863684}{:}
\colorize{39.102669625535896}{ PEEK}
\colorize{16.43010757737003}{ is}
\colorize{12.496428665103446}{ a highly durable material}
\colorize{10.207280886089128}{\sdelete{ that}}
\colorize{21.71537418195135}{ exhibits}
\colorize{7.31976222944261}{\sdelete{ excellent mechanical properties}}
\colorize{12.05202246833176}{,}
\colorize{9.4259481286046}{\sdelete{ including}}
\colorize{3.581351946074597}{\sdelete{ high tensile strength}}
\colorize{3.197628450137016}{\sdelete{,}}
\colorize{23.473082143345124}{ stiffness}
\colorize{3.0171539062824015}{\sdelete{,}}
\colorize{5.870108352475634}{\sdelete{ and}}
\colorize{16.903186196037794}{ toughness}
\colorize{1.895552662898659}{\sdelete{.}}
\colorize{25.72184062198061}{This}
\colorize{47.46806638181497}{ makes}
\colorize{17.718882653029983}{ it}
\colorize{43.80154879721664}{ ideal}
\colorize{1.0909031304606749}{\sdelete{ for}}
\colorize{18.37824905578323}{ use}
\colorize{5.929053838912792}{\sdelete{ in}}
\colorize{29.846646912109815}{ applications}
\colorize{4.431702761862224}{\sdelete{ where}}
\colorize{30.34773490933762}{ the membrane}
\colorize{23.14323520450618}{ may}
\colorize{2.8777409825235933}{\sdelete{ be}}
\colorize{17.71050776199119}{ subjected}
\colorize{0.028490819369198134}{\sdelete{ to}}
\colorize{13.74973341258926}{ mechanical stress}
\colorize{11.320785280832084}{ or}
\colorize{16.23417148549504}{ deformation}
\colorize{5.806274400053001}{\sdelete{.}}
\colorize{26.225223900916504}{Biocompatibility}
\colorize{24.59495333564014}{:}
\colorize{29.840507068560562}{ PEEK}
\colorize{10.999097821769706}{ is}
\colorize{2.687341432830361}{\sdelete{ a biocompatible material}}
\colorize{11.575012905416383}{ that}
\colorize{6.230916613515747}{\sdelete{ is}}
\colorize{27.017209399022658}{ well}
\colorize{11.465410285530421}{-}
\colorize{1.0696596158572083}{\sdelete{tolerated}}
\colorize{2.217849997315007}{\sdelete{ by}}
\colorize{3.5715842898360792}{\sdelete{ the human body}}
\colorize{11.20035357570016}{ and}
\colorize{13.18206123389007}{ does}
\colorize{0.009987539811051464}{\sdelete{ not}}
\colorize{8.789768320183626}{\sdelete{ cause}}
\colorize{11.546105772405111}{ adverse reactions}
\colorize{18.549095712337927}{ or}
\colorize{39.294360745878116}{ rejection}
\colorize{4.926492469973617}{\sdelete{.}}
\colorize{25.721846847739638}{This}
\colorize{47.46809751061009}{ makes}
\colorize{17.71887393696735}{ it}
\colorize{43.80158926465029}{ ideal}
\colorize{1.0908978385655046}{\sdelete{ for}}
\colorize{18.37820547547006}{ use}
\colorize{5.929039519667039}{\sdelete{ in}}
\colorize{35.781726180934434}{ biomedical applications}
\colorize{8.008621090318007}{\sdelete{,}}
\colorize{17.737898611392218}{ including}
\colorize{16.77927972877446}{ implantable devices}
\colorize{10.148184113707694}{\sdelete{.}}
\colorize{23.688322492918353}{Hydrophobicity}
\colorize{18.807698800364133}{:}
\colorize{44.07271702339259}{ PEEK}
\colorize{9.360189794067718}{\sdelete{ is}}
\colorize{5.445733348341158}{\sdelete{ a hydrophobic material}}
\colorize{15.252779191418883}{ that}
\colorize{22.2030310379492}{ resists}
\colorize{13.800697164669984}{ the absorption}
\colorize{0.29661504160534385}{\sdelete{ of}}
\colorize{4.564075472801759}{\sdelete{ water}}
\colorize{10.134631881464726}{\sdelete{ and}}
\colorize{9.260994775541814}{\sdelete{ other liquids}}
\colorize{3.279566287224902}{\sdelete{.}}
\colorize{33.9268898417227}{This property}
\colorize{42.59907457056843}{ makes}
\colorize{15.889833718982699}{ it}
\colorize{37.26212794033071}{ ideal}
\colorize{0.8187034205296586}{\sdelete{ for}}
\colorize{27.429726324210684}{ use}
\colorize{14.093595157566641}{ in}
\colorize{46.06579509657559}{ applications}
\colorize{5.553168470472019}{\sdelete{ where}}
\colorize{34.11194741582508}{ the membrane}
\colorize{17.56263540138605}{ needs}
\colorize{0.21263551665168154}{\sdelete{ to}}
\colorize{19.704850598119865}{ remain}
\colorize{18.075111868308106}{ dry}
\colorize{19.333416211577966}{ or}
\colorize{37.46786001499927}{ maintain}
\colorize{20.673734034907852}{ its structure}
\colorize{17.571282980670087}{ in}
\colorize{9.406670688363365}{\sdelete{ the presence}}
\colorize{0.02123608432609176}{\sdelete{ of}}
\colorize{17.5501346998422}{ liquids}
\colorize{3.0935175722859354}{\sdelete{.}}
\colorize{51.61521878880823}{Overall}
\colorize{10.874567699472426}{\sdelete{,}}
\colorize{29.420437675100786}{ the combination}
\colorize{0.39907580693020855}{\sdelete{ of}}
\colorize{31.479705216633057}{ these properties}
\colorize{12.855236901057168}{ makes}
\colorize{21.503694963426845}{ PEEK an ideal material}
\colorize{0.8178384135708945}{\sdelete{ for}}
\colorize{45.930604295813936}{ membrane manufacturing}
\colorize{12.558773727255543}{,}
\colorize{15.967356870346821}{ particularly}
\colorize{9.602838129443684}{\sdelete{ in}}
\colorize{15.211143182795542}{ applications}
\colorize{4.590664444440524}{\sdelete{ where}}
\colorize{30.30481047988431}{ durability}
\colorize{7.3620872220645825}{\sdelete{,}}
\colorize{8.917849833329637}{\sdelete{ chemical resistance}}
\colorize{1.9941035021383724}{\sdelete{,}}
\colorize{4.7050733502928415}{\sdelete{ and}}
\colorize{14.71845653608238}{ thermal stability}
\colorize{0.7035590657638613}{\sdelete{ are}}
\colorize{10.914088195179486}{ critical}
\colorize{1.633317467057975}{\sdelete{.}}

\textbf{human:}
\colorize{36.46602939106675}{2 / 2Membrane Manufacturing Processes}
\colorize{46.38838902615183}{ using}
\colorize{25.13549770325706}{ PEEK}

\textbf{gpt:}
\colorize{31.625919035034265}{There}
\colorize{6.28068719887956}{\sdelete{ are}}
\colorize{43.141132575848005}{ several membrane manufacturing processes}
\colorize{9.293256036227145}{\sdelete{ that}}
\colorize{21.13638979910595}{ use}
\colorize{28.78946540281525}{ PEEK}
\colorize{11.760596556156273}{,}
\colorize{11.283237728159662}{ including}
\colorize{14.905547443081929}{:}
\colorize{39.310582168539874}{ Phase inversion}
\colorize{29.686632453680595}{:}
\colorize{16.497137834475474}{ This}
\colorize{8.981880924411758}{\sdelete{ is}}
\colorize{10.108368880192183}{\sdelete{ a widely used membrane manufacturing process}}
\colorize{8.995794873254006}{\sdelete{ that}}
\colorize{13.204369373624084}{ involves}
\colorize{14.678028947285556}{ the formation}
\colorize{0.22034303122615864}{\sdelete{ of}}
\colorize{14.463884419498768}{ a polymer solution}
\colorize{13.43895223818474}{,}
\colorize{9.413736302279455}{\sdelete{ followed}}
\colorize{0.03160598996040944}{\sdelete{ by}}
\colorize{39.6487209882973}{ casting}
\colorize{20.76309204288823}{ or}
\colorize{2.618894075096735}{\sdelete{ extrusion}}
\colorize{16.482535939261098}{ into}
\colorize{12.277881355273797}{ a flat or tubular membrane shape}
\colorize{2.788768162254016}{\sdelete{.}}
\colorize{45.62810649379056}{The membrane}
\colorize{17.175248752497517}{ is}
\colorize{36.65946014758295}{ then}
\colorize{29.875683852196733}{ immersed}
\colorize{1.068804612032993}{\sdelete{ in}}
\colorize{19.613337618362515}{ a nonsolvent bath}
\colorize{15.824720995656582}{ to}
\colorize{37.1307040350664}{ induce}
\colorize{10.483192486936508}{\sdelete{ phase separation}}
\colorize{13.293260138389728}{,}
\colorize{20.586730608990266}{ resulting}
\colorize{0.04841797359639671}{\sdelete{ in}}
\colorize{18.17920718176163}{ a porous structure}
\colorize{8.170907328215007}{\sdelete{.}}
\colorize{43.47965339780286}{PEEK}
\colorize{55.27699857078946}{ can}
\colorize{4.407047510976246}{\sdelete{ be}}
\colorize{8.347604953100255}{\sdelete{ used}}
\colorize{9.608905754188262}{\sdelete{ in}}
\colorize{32.27434740175863}{ this process}
\colorize{20.326994432821728}{ by}
\colorize{15.253505177178647}{ dissolving}
\colorize{6.3212853734736365}{\sdelete{ it}}
\colorize{1.9811400399262844}{\sdelete{ in}}
\colorize{7.573206134728464}{\sdelete{ a suitable solvent}}
\colorize{11.297713240465855}{ and}
\colorize{9.974209635819447}{\sdelete{ then}}
\colorize{27.115205336360575}{ casting}
\colorize{22.874802888133228}{ or}
\colorize{7.298462072260968}{\sdelete{ extruding}}
\colorize{2.7958411848233435}{\sdelete{ it}}
\colorize{6.1065085160953325}{\sdelete{ into}}
\colorize{22.413144386768085}{ a membrane shape}
\colorize{3.9930046522514884}{\sdelete{.}}
\colorize{25.79445118254305}{Electrospinning}
\colorize{22.120743935780077}{:}
\colorize{24.93955750238111}{ This process}
\colorize{12.264967707405264}{ involves}
\colorize{8.520296613341651}{\sdelete{ the use}}
\colorize{0.055816024367078734}{\sdelete{ of}}
\colorize{7.649074837325208}{\sdelete{ an electrostatic field}}
\colorize{2.814374273315696}{\sdelete{ to}}
\colorize{11.009105106824531}{ draw}
\colorize{17.381997493614655}{ out}
\colorize{12.155087211141545}{ a polymer solution}
\colorize{10.23597603200564}{\sdelete{ into}}
\colorize{6.579624339899021}{\sdelete{ nanofibers}}
\colorize{15.410003264653643}{,}
\colorize{24.541514533059516}{ resulting}
\colorize{0.05256067458568797}{\sdelete{ in}}
\colorize{19.30879918685239}{ a highly porous membrane structure}
\colorize{3.006641649913157}{\sdelete{.}}
\colorize{43.47965339780286}{PEEK}
\colorize{55.27699857078946}{ can}
\colorize{4.407047510976246}{\sdelete{ be}}
\colorize{8.347604953100255}{\sdelete{ used}}
\colorize{9.608905754188262}{\sdelete{ in}}
\colorize{32.27434740175863}{ this process}
\colorize{20.326994432821728}{ by}
\colorize{15.253505177178647}{ dissolving}
\colorize{6.3212853734736365}{\sdelete{ it}}
\colorize{1.9811400399262844}{\sdelete{ in}}
\colorize{7.573206134728464}{\sdelete{ a suitable solvent}}
\colorize{11.297713240465855}{ and}
\colorize{9.974209635819447}{\sdelete{ then}}
\colorize{23.246668090973817}{ electrospinning}
\colorize{4.914286246831591}{\sdelete{ it}}
\colorize{9.49310850407364}{\sdelete{ into}}
\colorize{11.2609096161907}{ a nanofiber membrane}
\colorize{3.7178630221513744}{\sdelete{.}}
\colorize{30.33730520653303}{Thermally}
\colorize{17.822670413987733}{ induced}
\colorize{17.663975816471552}{ phase}
\colorize{8.004861977019448}{\sdelete{ separation}}
\colorize{30.123789687932945}{:}
\colorize{31.422052197769233}{ This process}
\colorize{19.38125681164441}{ involves}
\colorize{9.438554668051856}{\sdelete{ the formation}}
\colorize{0.21990568410625608}{\sdelete{ of}}
\colorize{22.688431886402128}{ a polymer solution}
\colorize{19.399094856399344}{,}
\colorize{7.868230846907647}{\sdelete{ which}}
\colorize{7.802428932057596}{\sdelete{ is}}
\colorize{13.011923691288182}{ then}
\colorize{17.480330244515766}{ heated}
\colorize{6.1652952456773855}{\sdelete{ to}}
\colorize{15.406296447730886}{ induce}
\colorize{3.7891847323064622}{\sdelete{ phase separation}}
\colorize{19.874298326621922}{ and}
\colorize{30.090945696202986}{ create}
\colorize{21.980601586529648}{ a porous membrane structure}
\colorize{1.9805786009775201}{\sdelete{.}}
\colorize{43.47965339780286}{PEEK}
\colorize{55.27699857078946}{ can}
\colorize{4.407047510976246}{\sdelete{ be}}
\colorize{8.347604953100255}{\sdelete{ used}}
\colorize{9.608905754188262}{\sdelete{ in}}
\colorize{32.27434740175863}{ this process}
\colorize{20.326994432821728}{ by}
\colorize{15.253505177178647}{ dissolving}
\colorize{6.3212853734736365}{\sdelete{ it}}
\colorize{1.9811400399262844}{\sdelete{ in}}
\colorize{7.573206134728464}{\sdelete{ a suitable solvent}}
\colorize{11.297713240465855}{ and}
\colorize{9.974209635819447}{\sdelete{ then}}
\colorize{23.522467976801273}{ heating}
\colorize{7.196614018123309}{\sdelete{ it}}
\colorize{5.204431520375887}{\sdelete{ to}}
\colorize{37.0882692615603}{ induce}
\colorize{9.8589425090857}{\sdelete{ phase separation}}
\colorize{3.351414659238848}{\sdelete{.}}
\colorize{41.808418116369886}{Polymer}
\colorize{58.34401914171083}{ blending}
\colorize{25.263733086037337}{:}
\colorize{22.90379562533107}{ This process}
\colorize{11.689861349855779}{ involves}
\colorize{15.227525022514815}{ blending}
\colorize{35.59574314775317}{ PEEK}
\colorize{9.067917801240899}{\sdelete{ with}}
\colorize{5.520461109249598}{\sdelete{ other polymers}}
\colorize{9.411889742153}{\sdelete{ to}}
\colorize{10.132545007039955}{\sdelete{ create}}
\colorize{31.413954975582904}{ a membrane}
\colorize{7.781124384678217}{\sdelete{ with}}
\colorize{13.046543394006902}{ specific properties}
\colorize{2.8585692556705813}{\sdelete{.}}
\colorize{33.88052412396899}{For}
\colorize{33.0561626555675}{ example}
\colorize{1.8073027155093322}{\sdelete{,}}
\colorize{60.83049702230316}{ blending PEEK}
\colorize{11.649361542254278}{ with}
\colorize{9.771479270396775}{\sdelete{ polyvinylidene fluoride}}
\colorize{2.4475521158313813}{\sdelete{ (}}
\colorize{0.2871727338056612}{\sdelete{PVDF}}
\colorize{1.1761396859673108}{\sdelete{)}}
\colorize{17.349713197620932}{ can}
\colorize{20.624235826501327}{ create}
\colorize{21.381074268970643}{ a membrane}
\colorize{6.340981807297173}{\sdelete{ with}}
\colorize{12.9615991171339}{ enhanced mechanical strength and chemical resistance}
\colorize{3.194135799324676}{\sdelete{.}}
\colorize{51.61521878880823}{Overall}
\colorize{10.874567699472426}{\sdelete{,}}
\colorize{49.129741698978975}{ PEEK}
\colorize{29.767071127995553}{ can}
\colorize{3.0184217198500045}{\sdelete{ be}}
\colorize{9.15239388028278}{\sdelete{ used}}
\colorize{8.012868925699934}{\sdelete{ in}}
\colorize{8.872487801316948}{\sdelete{ a variety}}
\colorize{0.010708644575739931}{\sdelete{ of}}
\colorize{36.435956726791126}{ membrane manufacturing processes}
\colorize{19.637853960113155}{ to}
\colorize{16.079371971854496}{ create}
\colorize{17.798051272504093}{ membranes}
\colorize{6.456326444730765}{\sdelete{ with}}
\colorize{12.835260152433586}{ specific properties}
\colorize{12.66741508994823}{,}
\colorize{8.513360806501206}{\sdelete{ including}}
\colorize{16.893511898817234}{ porosity}
\colorize{2.1399559152282484}{\sdelete{,}}
\colorize{11.185349496452849}{ mechanical strength}
\colorize{1.1108648438762143}{\sdelete{,}}
\colorize{6.323174268761451}{\sdelete{ and}}
\colorize{8.25893286812032}{\sdelete{ chemical resistance}}
\colorize{0.7102575155415228}{\sdelete{.}}
\colorize{37.74817761663657}{The specific manufacturing process}
\colorize{15.465703885487612}{ used}
\colorize{31.015881811330047}{ will}
\colorize{9.368302580651607}{\sdelete{ depend}}
\colorize{1.0431612086747608}{\sdelete{ on}}
\colorize{14.70158638933034}{ the desired properties}
\colorize{2.6449117897647545}{\sdelete{ of}}
\colorize{22.11357722798896}{ the membrane}
\colorize{13.26415720525716}{ and}
\colorize{9.422238302564981}{\sdelete{ the intended application}}
\colorize{1.8401085433364737}{\sdelete{.}}

%% file: text_3.tex
\colorize{60.457407703812116}{University}
\colorize{2.1659164753380376}{\sdelete{ of}}
\colorize{39.18128312793362}{ Edinburgh}
\colorize{40.9589778094108}{ student}
\colorize{42.183778727475435}{ Romy}
\colorize{40.15700491764415}{ Ulvestad}
\colorize{39.06025167513745}{ died}
\colorize{26.786787725140492}{ at}
\colorize{20.085639942990706}{ her parents' London home}
\colorize{41.68918976805701}{ four}
\colorize{10.03597654942884}{\sdelete{ days}}
\colorize{2.3817889494236595}{\sdelete{ after}}
\colorize{9.6403311919711}{\sdelete{ her 21st birthday}}
\colorize{9.46694401856072}{\sdelete{.}}
\colorize{45.47825276865044}{She}
\colorize{29.067922211898566}{ had}
\colorize{56.20448609818468}{ killed}
\colorize{20.441726675655065}{ herself}
\colorize{23.055039819628853}{ after}
\colorize{68.70521097319349}{ staff}
\colorize{21.92413100454727}{ failed}
\colorize{0.09019493549262621}{\sdelete{ to}}
\colorize{22.57557129376705}{ provide}
\colorize{36.01242053741374}{ additional support}
\colorize{14.840490912826997}{\sdelete{,}}
\colorize{29.86040639768425}{ despite}
\colorize{20.117131135206947}{ repeated warnings}
\colorize{14.326312443816134}{\sdelete{ that}}
\colorize{2.808450243096371}{\sdelete{ she}}
\colorize{1.9242394345653322}{\sdelete{ was}}
\colorize{22.97116717758698}{ struggling}
\colorize{12.349226839923665}{\sdelete{ with}}
\colorize{4.373718934658043}{\sdelete{ her mental health}}
\colorize{3.9098697400883897}{\sdelete{.}}
\colorize{47.210863096203305}{Now}
\colorize{44.18451179062587}{ her mother}
\colorize{15.050620498972265}{\sdelete{ is}}
\colorize{50.434400842414064}{ adding}
\colorize{19.393786310850537}{ her support}
\colorize{9.24085743362491}{\sdelete{ to}}
\colorize{17.927550447837522}{ a campaign}
\colorize{19.32284822682462}{ for}
\colorize{24.498929706108818}{ the UK government}
\colorize{0.3057201169280982}{\sdelete{ to}}
\colorize{33.128338635641825}{ create}
\colorize{45.486541491338194}{ legal duty}
\colorize{14.68116423730779}{\sdelete{ of}}
\colorize{0.17729261165188498}{\sdelete{ care}}
\colorize{3.2233977628699466}{\sdelete{ for}}
\colorize{57.40666397369997}{ students}
\colorize{12.786306054351288}{\sdelete{.}}
\colorize{41.1834574591603}{The university}
\colorize{29.82275380591079}{,}
\colorize{8.119198782969134}{\sdelete{ which}}
\colorize{65.79097980426549}{ apologised}
\colorize{7.5368563489480245}{\sdelete{ for}}
\colorize{48.01010277883098}{ failing Romy}
\colorize{9.946663728680099}{\sdelete{,}}
\colorize{9.075593935880999}{\sdelete{ said}}
\colorize{8.122517134542898}{\sdelete{ it}}
\colorize{8.895369588880868}{\sdelete{ had}}
\colorize{38.2603836740535}{ improved}
\colorize{21.132530615223434}{ the support}
\colorize{7.906576676114055}{\sdelete{ it}}
\colorize{10.063074393011352}{\sdelete{ offers}}
\colorize{49.95147584941807}{ since}
\colorize{6.278278376048224}{\sdelete{ her death}}
\colorize{4.97138557635926}{\sdelete{.}}
\colorize{44.09813839819054}{Libby Kitson}
\colorize{15.394068097795063}{\sdelete{,}}
\colorize{34.693620172666236}{ Romy's mother}
\colorize{1.7575610037328628}{\sdelete{,}}
\colorize{17.261966552729486}{\sdelete{ has}}
\colorize{43.354710283222495}{ joined}
\colorize{35.83918927464454}{ calls}
\colorize{1.1529112607648009}{\sdelete{ for}}
\colorize{55.580556567979386}{ Westminster}
\colorize{5.766183090438528}{\sdelete{ to}}
\colorize{29.63479785002497}{ change}
\colorize{2.880443157879186}{\sdelete{ the law}}
\colorize{8.99467324558417}{\sdelete{ to}}
\colorize{38.21027563497761}{ better}
\colorize{0.7660154049091029}{\sdelete{ protect}}
\colorize{48.32643607470396}{ students}
\colorize{26.07953229013774}{,}
\colorize{23.338970346274227}{ and}
\colorize{36.08285589511217}{ hopes}
\colorize{24.53940329031355}{ the Scottish government}
\colorize{2.099177748411536}{\sdelete{ will}}
\colorize{10.326601744819643}{\sdelete{ follow}}
\colorize{1.7193631059643177}{\sdelete{ suit}}
\colorize{2.0213975053966364}{\sdelete{.}}
\colorize{45.62717891309411}{The petition}
\colorize{14.978293745283006}{\sdelete{ to}}
\colorize{19.77630164922276}{ the UK government}
\colorize{24.26539871631973}{ is}
\colorize{22.201993034029503}{ being}
\colorize{20.950746922077258}{ supported}
\colorize{0.19049536432371533}{\sdelete{ by}}
\colorize{31.600863716076667}{ 25 bereaved families}
\colorize{10.316027674762726}{\sdelete{,}}
\colorize{15.915616069610891}{\sdelete{ who}}
\colorize{10.592864929724009}{\sdelete{ have}}
\colorize{32.32725407262657}{ come}
\colorize{1.4271352691352037}{\sdelete{ together}}
\colorize{31.134903870642127}{ and}
\colorize{23.62105026753471}{ set}
\colorize{0.8539278555312274}{\sdelete{ up}}
\colorize{34.64145811905055}{ The LEARN Network}
\colorize{8.89207270604707}{\sdelete{.}}
\colorize{33.59967766530393}{It}
\colorize{76.41553002508377}{ reached}
\colorize{22.84302189201738}{ more than 127,000 signatures}
\colorize{20.014493025374914}{ before}
\colorize{15.426292676480966}{\sdelete{ the deadline}}
\colorize{31.18032242128522}{ at}
\colorize{9.557527788715326}{\sdelete{ midnight}}
\colorize{7.44278462417394}{\sdelete{ on}}
\colorize{31.135203025606444}{ Mothers' Day}
\colorize{9.794542677917459}{\sdelete{,}}
\colorize{30.002163056416958}{ meaning}
\colorize{38.750253804427835}{ Parliament}
\colorize{5.977435348558973}{\sdelete{ will}}
\colorize{23.787175377375842}{ consider}
\colorize{28.409173605782716}{ the subject}
\colorize{19.87493762811318}{ for}
\colorize{4.872126574635082}{\sdelete{ debate}}
\colorize{5.763261481179298}{\sdelete{.}}
\colorize{58.946230253995545}{Speaking}
\colorize{11.14792417337425}{\sdelete{ to}}
\colorize{29.87777961768776}{ BBC Scotland}
\colorize{6.3793517009867555}{\sdelete{,}}
\colorize{36.876629529201324}{ Ms Kitson}
\colorize{3.503621353748212}{\sdelete{ said}}
\colorize{15.716555765791046}{\sdelete{ she}}
\colorize{27.295362853020766}{ did}
\colorize{0.37979150630178377}{\sdelete{ not}}
\colorize{15.788561919630173}{\sdelete{ know}}
\colorize{31.42550473479431}{ that}
\colorize{17.665644343708845}{\sdelete{ her daughter}}
\colorize{7.308104062133356}{\sdelete{ was}}
\colorize{39.979056495621755}{ struggling}
\colorize{5.382328746754758}{\sdelete{ with}}
\colorize{7.275225144878336}{\sdelete{ her mental health}}
\colorize{25.60600845837196}{ when}
\colorize{1.3810954847200225}{\sdelete{ she}}
\colorize{29.40772227180149}{ returned}
\colorize{5.60480572087796}{\sdelete{ to}}
\colorize{43.07544992314394}{ London}
\colorize{47.31381572307581}{ during}
\colorize{90.10896237769124}{ the Covid lockdown}
\colorize{19.971360895537984}{ in}
\colorize{84.636211119644}{ 2020}
\colorize{1.267641495256607}{\sdelete{.}}
\colorize{34.986746476262844}{"}
\colorize{42.455898584641965}{Her father}
\colorize{26.383694957325154}{ and}
\colorize{11.989613279908003}{\sdelete{ I}}
\colorize{21.194082923949445}{ had}
\colorize{16.969137376493983}{\sdelete{ no idea}}
\colorize{27.106437225713243}{ about}
\colorize{20.337161209692763}{ her mental health struggles}
\colorize{35.54375365025248}{ but}
\colorize{31.64066945608579}{ they}
\colorize{26.984856173174006}{ had}
\colorize{13.980090368165325}{\sdelete{ been}}
\colorize{36.862328593230366}{ well}
\colorize{14.602431927237951}{\sdelete{-}}
\colorize{44.40050091153196}{flagged}
\colorize{18.32649545509183}{ to}
\colorize{47.2595949890475}{ people}
\colorize{22.969583500202305}{ at}
\colorize{21.04745071437092}{ the University}
\colorize{8.192289108176439}{\sdelete{ of}}
\colorize{34.68673514775979}{ Edinburgh}
\colorize{24.19710665449437}{ who}
\colorize{27.02675671296167}{ could}
\colorize{9.776433080133192}{\sdelete{ have}}
\colorize{14.636265945796492}{\sdelete{ done}}
\colorize{1.853041077848584}{\sdelete{ something}}
\colorize{5.779948417128621}{\sdelete{ about}}
\colorize{1.8275691335060091}{\sdelete{ it}}
\colorize{4.780494128095702}{\sdelete{,}}
\colorize{4.780494128095702}{\sdelete{"}}
\colorize{15.752506602110028}{\sdelete{ she}}
\colorize{3.151130978527658}{\sdelete{ said}}
\colorize{0.8942890866634745}{\sdelete{.}}
\colorize{34.986746476262844}{"}
\colorize{16.633012621747955}{\sdelete{I}}
\colorize{54.39789163021735}{ sent}
\colorize{45.876570173073134}{ Romy}
\colorize{23.026756185975664}{ to}
\colorize{71.39069987405212}{ off}
\colorize{14.557486404498738}{\sdelete{ to}}
\colorize{37.861673391689465}{ university}
\colorize{18.54854162316642}{ to}
\colorize{12.608348090666688}{\sdelete{ study}}
\colorize{63.589024177368515}{ classics}
\colorize{10.204955570435846}{\sdelete{,}}
\colorize{24.16981272975352}{ which}
\colorize{33.947204324719884}{ would}
\colorize{15.345083589443956}{\sdelete{ be}}
\colorize{33.305721714178496}{ a pivotal point}
\colorize{17.52648576032411}{\sdelete{ for}}
\colorize{13.209261700381287}{\sdelete{ the rest}}
\colorize{0.0022580176631895434}{\sdelete{ of}}
\colorize{2.412075079796986}{\sdelete{ her life}}
\colorize{3.422051019380074}{\sdelete{.}}
\colorize{38.550912975739124}{What}
\colorize{26.504642682035456}{ I}
\colorize{32.50175700042631}{ did}
\colorize{3.2780800748074865}{\sdelete{n't}}
\colorize{18.448689082344707}{ expect}
\colorize{14.077272197735521}{\sdelete{ to}}
\colorize{23.457905018800577}{ do}
\colorize{11.275812026090236}{\sdelete{ was}}
\colorize{51.735269084294345}{ send}
\colorize{34.58484236101684}{ her}
\colorize{21.905221338387012}{ off}
\colorize{7.3029451239200185}{\sdelete{ to}}
\colorize{35.32171392279987}{ university}
\colorize{27.508026466106443}{ to}
\colorize{61.741982603309076}{ die}
\colorize{16.26989250087151}{\sdelete{.}}
\colorize{16.26989250087151}{\sdelete{"}}
\colorize{42.77412864369076}{ Romy}
\colorize{24.346667200203974}{,}
\colorize{31.816652772776816}{ a classics student}
\colorize{22.409162796558398}{ and}
\colorize{11.965833125946437}{\sdelete{ part-time model}}
\colorize{1.5558488095902034}{\sdelete{,}}
\colorize{34.06688038439965}{ died}
\colorize{8.914302155030422}{\sdelete{ in}}
\colorize{28.818529878482547}{ April}
\colorize{69.59632827537094}{ 2020}
\colorize{9.2515481505023}{\sdelete{,}}
\colorize{46.92341384307783}{ more}
\colorize{0.07167278724496537}{\sdelete{ than}}
\colorize{7.726749500803258}{\sdelete{ a}}
\colorize{5.327463950527524}{\sdelete{ year}}
\colorize{2.857320824000734}{\sdelete{ after}}
\colorize{8.853670139624201}{\sdelete{ she}}
\colorize{92.95829622930067}{ communicated}
\colorize{8.008674131511334}{\sdelete{ with}}
\colorize{26.032100329498363}{ the university}
\colorize{13.006871594993033}{\sdelete{ about}}
\colorize{23.60635133690605}{ her struggles}
\colorize{63.89360834748389}{ by}
\colorize{50.82739900869835}{ requesting}
\colorize{20.788948073442885}{ a "special circumstances" application}
\colorize{15.433899051050199}{\sdelete{ to}}
\colorize{28.2773959751976}{ resit}
\colorize{8.0386182290758}{\sdelete{ her exams}}
\colorize{37.43075713688083}{ without}
\colorize{23.276054772560506}{ penalty}
\colorize{1.031273602314523}{\sdelete{.}}
\colorize{53.60545620232554}{Ms Kitson}
\colorize{13.28059302016088}{\sdelete{ said}}
\colorize{17.840384501085182}{:}
\colorize{3.751304918588034}{\sdelete{ "}}
\colorize{17.06770522791625}{\sdelete{We}}
\colorize{27.233633069377728}{ want}
\colorize{47.94264942911355}{ there}
\colorize{0.04729730154607699}{\sdelete{ to}}
\colorize{0.016603111254132264}{\sdelete{ be}}
\colorize{25.720108132116877}{ a legal duty}
\colorize{6.466238625682035}{\sdelete{ of}}
\colorize{0.3716672484746483}{\sdelete{ care}}
\colorize{43.474528752210695}{ within}
\colorize{21.29369390566533}{ all higher education institutions}
\colorize{14.646487212893334}{\sdelete{,}}
\colorize{14.815567136978169}{\sdelete{ and}}
\colorize{25.707411042219075}{ I}
\colorize{8.097491024366906}{\sdelete{ think}}
\colorize{31.411865644295204}{ some people}
\colorize{32.03279956965087}{,}
\colorize{11.951830444445427}{\sdelete{ including}}
\colorize{28.787486438306992}{ our government}
\colorize{1.2190530135953708}{\sdelete{,}}
\colorize{41.56405778592773}{ seem}
\colorize{0.6457245520792312}{\sdelete{ to}}
\colorize{6.002818984817829}{\sdelete{ think}}
\colorize{17.619752407349548}{\sdelete{ there}}
\colorize{6.9387103460646475}{\sdelete{ is}}
\colorize{30.506586368529753}{ some legal requirement}
\colorize{26.42844583015536}{ in}
\colorize{14.203129107650916}{\sdelete{ place}}
\colorize{23.61528805772838}{ but}
\colorize{9.68781155281129}{\sdelete{ there}}
\colorize{25.234779594411133}{ really}
\colorize{0.7526871459834433}{\sdelete{ is}}
\colorize{0.17517059845636715}{\sdelete{n't}}
\colorize{7.023881560624473}{\sdelete{.}}
\colorize{34.986746476262844}{"}
\colorize{29.704582704902492}{You}
\colorize{28.952926337530943}{ want}
\colorize{4.833778109436924}{\sdelete{ to}}
\colorize{13.331113259044072}{\sdelete{ know}}
\colorize{45.59050565367534}{ that}
\colorize{30.0009686788503}{ if}
\colorize{19.84446203610967}{ your child}
\colorize{7.158977184971025}{\sdelete{ is}}
\colorize{43.6287537394898}{ suffering}
\colorize{26.513078465591967}{ with}
\colorize{17.93113250710051}{ any kind}
\colorize{0.0}{\sdelete{ of}}
\colorize{14.616931914209614}{\sdelete{ mental health issues}}
\colorize{4.396616168797972}{\sdelete{,}}
\colorize{24.842936446336907}{ or}
\colorize{21.895683492845688}{ is}
\colorize{56.28235035553531}{ overwhelmed}
\colorize{9.92157893728154}{\sdelete{ by}}
\colorize{35.016676404458934}{ university life}
\colorize{1.591856824319958}{\sdelete{,}}
\colorize{23.95090099771367}{ there}
\colorize{12.088697955463672}{\sdelete{ are}}
\colorize{8.307510404313687}{\sdelete{ people}}
\colorize{20.66653300806436}{ there}
\colorize{15.989070647772918}{\sdelete{ who}}
\colorize{10.834771113514595}{\sdelete{ are}}
\colorize{17.399489575941296}{\sdelete{ trained}}
\colorize{13.928015792508738}{\sdelete{ and}}
\colorize{28.75901781567778}{ skilled}
\colorize{4.699179844232869}{\sdelete{ to}}
\colorize{3.8161598351451644}{\sdelete{ help}}
\colorize{53.00779071755086}{ deal}
\colorize{0.014164490678494554}{\sdelete{ with}}
\colorize{13.075331828614779}{\sdelete{ it}}
\colorize{9.525681059852788}{\sdelete{.}}
\colorize{34.98308534047432}{"}
\colorize{40.8372693756811}{And}
\colorize{29.57097557986011}{ if}
\colorize{23.64552391546896}{ they}
\colorize{44.3569333240861}{ really}
\colorize{31.503050539168804}{,}
\colorize{2.0091309830058943}{\sdelete{ really}}
\colorize{30.71663341929693}{ are}
\colorize{29.8569034149468}{ in}
\colorize{21.671145161873618}{ a state}
\colorize{1.1565116400192237}{\sdelete{ of}}
\colorize{28.481562261005532}{ crisis}
\colorize{45.972894684294104}{ you}
\colorize{26.151620740813396}{ would}
\colorize{17.76410606799323}{\sdelete{ want}}
\colorize{3.6726924756496646}{\sdelete{ to}}
\colorize{19.198300194422362}{ be}
\colorize{47.36880432237018}{ informed}
\colorize{10.625185044302754}{\sdelete{.}}
\colorize{10.625185044302754}{\sdelete{"}}
\colorize{51.04761457309744}{ Libby Kitson}
\colorize{60.18495508117614}{ hopes}
\colorize{23.25486953923485}{ the Scottish government}
\colorize{5.019378944548696}{\sdelete{ will}}
\colorize{23.685620852457568}{ legislate}
\colorize{40.54472415931286}{ after}
\colorize{36.415751751367615}{ the UK government Campaigners}
\colorize{27.135693393286797}{ from}
\colorize{55.54055648876165}{ ForThe100}
\colorize{37.619463067416056}{ want}
\colorize{21.09097679558619}{ a change}
\colorize{2.9118762556850872}{\sdelete{ in}}
\colorize{18.484974115382467}{ law}
\colorize{6.748960247580041}{\sdelete{ to}}
\colorize{15.81937886459911}{\sdelete{ ensure}}
\colorize{27.805634249899747}{ higher education institutions}
\colorize{20.71755489983804}{ have}
\colorize{13.354957634355824}{\sdelete{ a legal duty}}
\colorize{20.239463951457264}{ of}
\colorize{0.12650856489201442}{\sdelete{ care}}
\colorize{12.682314340464524}{\sdelete{ towards}}
\colorize{6.771665837434558}{\sdelete{ their students}}
\colorize{19.320193259315744}{,}
\colorize{23.323696769039202}{ as}
\colorize{37.37344991367036}{ schools}
\colorize{19.011892850312613}{ already}
\colorize{3.906118149342155}{\sdelete{ do}}
\colorize{2.1435645114119586}{\sdelete{.}}
\colorize{32.417658936614814}{The Department}
\colorize{0.637461556098202}{\sdelete{ of}}
\colorize{24.608738255818047}{ Education}
\colorize{12.315570296282164}{\sdelete{ has}}
\colorize{26.134062174021537}{ said}
\colorize{77.94128474933586}{ creating}
\colorize{32.35890602634579}{ a legally enforced duty}
\colorize{8.421026366873132}{\sdelete{ of}}
\colorize{2.842883955124569}{\sdelete{ care}}
\colorize{42.82185973544031}{ "}
\colorize{8.226246929802103}{\sdelete{would}}
\colorize{9.34754419000182}{\sdelete{ be}}
\colorize{16.008617458426997}{\sdelete{ a disproportionate response}}
\colorize{7.711641232865954}{\sdelete{"}}
\colorize{7.711641232865954}{\sdelete{.}}
\colorize{53.60545620232554}{Ms Kitson}
\colorize{13.28059302016088}{\sdelete{ said}}
\colorize{25.885162190375354}{ the campaign}
\colorize{9.962495493154188}{\sdelete{ was}}
\colorize{19.076997519877985}{ not}
\colorize{46.84592099492859}{ looking}
\colorize{4.988957024395097}{\sdelete{ to}}
\colorize{44.86294150629134}{ provide}
\colorize{28.668313462767042}{ a means}
\colorize{6.991693264109106}{\sdelete{ for}}
\colorize{45.315867513311034}{ families}
\colorize{1.0209361893344762}{\sdelete{ to}}
\colorize{25.75099473121991}{ sue}
\colorize{40.94168617061916}{ universities}
\colorize{11.797679166506862}{\sdelete{,}}
\colorize{3.3715428029912604}{\sdelete{ but}}
\colorize{10.270107471229144}{\sdelete{ to}}
\colorize{39.07642479257697}{ prevent}
\colorize{30.57086425306497}{ similar deaths}
\colorize{14.430042954700003}{\sdelete{.}}
\colorize{45.47825276865044}{She}
\colorize{29.526542444381366}{ said}
\colorize{13.335935853664644}{\sdelete{ she}}
\colorize{28.764478031875278}{ did}
\colorize{2.7008970083745174}{\sdelete{ not}}
\colorize{12.631374745527365}{\sdelete{ want}}
\colorize{32.69711227681148}{ any other parent}
\colorize{2.086286915062642}{\sdelete{ to}}
\colorize{9.319530346847511}{\sdelete{ have}}
\colorize{0.8214452941534216}{\sdelete{ to}}
\colorize{2.202830828706604}{\sdelete{ go}}
\colorize{0.03851444932994112}{\sdelete{ through}}
\colorize{13.571595826833494}{\sdelete{ the same situation}}
\colorize{6.826340527969411}{\sdelete{.}}
\colorize{34.986746476262844}{"}
\colorize{23.955086686381975}{We}
\colorize{36.04234438280018}{ would}
\colorize{40.613052717996155}{ hope}
\colorize{3.419499531197824}{\sdelete{ that}}
\colorize{26.858299859418892}{ the Scottish Parliament}
\colorize{4.60531171467611}{\sdelete{ would}}
\colorize{29.562287896421356}{ look}
\colorize{5.527346220838121}{\sdelete{ at}}
\colorize{54.892495617753745}{ England}
\colorize{10.935282744653204}{\sdelete{ and}}
\colorize{15.910535134486084}{\sdelete{ Wales}}
\colorize{8.439474455124165}{\sdelete{ and}}
\colorize{24.377577892004084}{ think}
\colorize{15.350743464923122}{\sdelete{ '}}
\colorize{36.81370261283666}{they}
\colorize{14.451347816615574}{\sdelete{'ve}}
\colorize{50.01602447861837}{ led}
\colorize{30.801611701446944}{ by}
\colorize{1.3850906741023825}{\sdelete{ example}}
\colorize{8.895803973052807}{\sdelete{,}}
\colorize{11.022516675748768}{\sdelete{ we}}
\colorize{10.800264407770523}{\sdelete{ should}}
\colorize{10.269051924847583}{\sdelete{ follow}}
\colorize{10.646860241982703}{\sdelete{ suit}}
\colorize{5.340410554892013}{\sdelete{'}}
\colorize{5.340410554892013}{\sdelete{,}}
\colorize{5.340410554892013}{\sdelete{"}}
\colorize{13.577271684596855}{\sdelete{ she}}
\colorize{2.404609933755484}{\sdelete{ said}}
\colorize{0.7944931517527984}{\sdelete{.}}
\colorize{34.986746476262844}{"}
\colorize{40.36398971505118}{A Scottish student's life}
\colorize{9.953950934648532}{\sdelete{ is}}
\colorize{49.757044205934456}{ no}
\colorize{22.382766265148042}{ less}
\colorize{12.374725978003527}{\sdelete{ important}}
\colorize{19.19868305361839}{.}
\colorize{19.19868305361839}{"}
\colorize{37.12342568367474}{ A}
\colorize{45.838324328452}{ University}
\colorize{1.5453859476063005}{\sdelete{ of}}
\colorize{8.698074015486062}{\sdelete{ Edinburgh}}
\colorize{29.88995382449665}{ spokeswoman}
\colorize{4.223967853881257}{\sdelete{ said}}
\colorize{21.76383823572968}{ it}
\colorize{12.829964168317334}{\sdelete{ had}}
\colorize{12.625876601222597}{\sdelete{ been}}
\colorize{13.689084460452491}{\sdelete{ "}}
\colorize{12.528455287858101}{\sdelete{shocked}}
\colorize{5.734196385391423}{\sdelete{ and}}
\colorize{19.733586012941124}{ deeply}
\colorize{0.845677418763077}{\sdelete{ saddened}}
\colorize{19.629887705167203}{ by}
\colorize{33.96448451351687}{ what}
\colorize{8.005264179980392}{\sdelete{ happened}}
\colorize{12.462455836992257}{\sdelete{"}}
\colorize{10.144370786257547}{\sdelete{ to}}
\colorize{48.67729647162399}{ Romy}
\colorize{21.105385540413003}{ and}
\colorize{25.77549557289145}{ "}
\colorize{7.13963596135626}{\sdelete{deeply}}
\colorize{19.432506256812545}{ sorry}
\colorize{5.090104631173904}{\sdelete{"}}
\colorize{3.4599765504009437}{\sdelete{ for}}
\colorize{46.96448747856456}{ the gaps}
\colorize{1.6180822498728011}{\sdelete{ in}}
\colorize{19.336229692444157}{ support}
\colorize{42.12590401390525}{ given}
\colorize{1.764063527255614}{\sdelete{ to}}
\colorize{4.4699342773362964}{\sdelete{ her}}
\colorize{3.980333436523572}{\sdelete{.}}
\colorize{34.986746476262844}{"}
\colorize{41.935250868655295}{Supporting}
\colorize{20.083970633958558}{ our students' mental health}
\colorize{13.820585502104413}{\sdelete{ and}}
\colorize{41.52544768835188}{ ensuring}
\colorize{13.182138809935484}{\sdelete{ their wellbeing}}
\colorize{2.747112335520137}{\sdelete{ is}}
\colorize{18.510693427317737}{ our absolute priority}
\colorize{6.059220585014149}{\sdelete{,}}
\colorize{6.059220585014149}{\sdelete{"}}
\colorize{19.110176026242957}{ she}
\colorize{1.4187475755562173}{\sdelete{ said}}
\colorize{1.0534049391562388}{\sdelete{.}}
\colorize{34.986746476262844}{"}
\colorize{52.60002062692204}{Our support policies}
\colorize{21.461678601980903}{ and}
\colorize{19.24655834373802}{ practices}
\colorize{6.283424719267451}{\sdelete{ are}}
\colorize{34.338581601136895}{ under}
\colorize{19.638750967117627}{ regular review}
\colorize{24.933663341424605}{ as}
\colorize{6.7216002706825675}{\sdelete{ we}}
\colorize{15.292550298376808}{\sdelete{ continue}}
\colorize{0.2775831720188374}{\sdelete{ to}}
\colorize{28.274445454701887}{ focus}
\colorize{37.322342862561484}{ and}
\colorize{11.65571211412251}{\sdelete{ improve}}
\colorize{21.729322226864955}{ upon}
\colorize{26.53145632243833}{ our mental health support provision}
\colorize{24.955375393774823}{ and}
\colorize{35.1340513730269}{ provide}
\colorize{13.82644462538106}{\sdelete{ the best possible environment}}
\colorize{1.3357487827933445}{\sdelete{ for}}
\colorize{7.68337012264441}{\sdelete{ our students}}
\colorize{5.128782569968044}{\sdelete{.}}
\colorize{5.128782569968044}{\sdelete{"}}
\colorize{60.142232288647854}{ Students}
\colorize{16.469251220176854}{\sdelete{ at}}
\colorize{15.590417257569792}{\sdelete{ the university}}
\colorize{43.769398317492374}{ now}
\colorize{2.59746605797561}{\sdelete{ have}}
\colorize{8.447787151238211}{\sdelete{ access}}
\colorize{0.052957198493982574}{\sdelete{ to}}
\colorize{36.10534106064265}{ a "named-contact relationship}
\colorize{5.033981051160116}{\sdelete{"}}
\colorize{5.383236874455232}{\sdelete{ with}}
\colorize{21.14560436223682}{ staff}
\colorize{35.74428170026119}{ trained}
\colorize{6.258525990281916}{\sdelete{ in}}
\colorize{41.51811468240486}{ wellbeing}
\colorize{10.28657900415416}{\sdelete{ and}}
\colorize{3.770121374635614}{\sdelete{ mental health}}
\colorize{13.927255083489348}{\sdelete{,}}
\colorize{41.09871783783622}{ alongside}
\colorize{12.726393241733861}{\sdelete{ other support services}}
\colorize{5.2816546688170884}{\sdelete{.}}
\colorize{51.76370514600127}{The spokeswoman}
\colorize{29.31194807594826}{ added}
\colorize{10.642527850257975}{\sdelete{:}}
\colorize{4.0005856145562415}{\sdelete{ "}}
\colorize{11.931037254165664}{\sdelete{We}}
\colorize{32.46087407891294}{ continue}
\colorize{0.16686169515304836}{\sdelete{ to}}
\colorize{31.091327337887947}{ engage}
\colorize{2.337808804358154}{\sdelete{ with}}
\colorize{59.32435201473415}{ students}
\colorize{28.21503429497495}{ through}
\colorize{24.68796534878158}{ both the Students' Association}
\colorize{1.7564730395539556}{\sdelete{ and}}
\colorize{23.292824005346358}{ the Sports Union}
\colorize{11.214059342457897}{\sdelete{ to}}
\colorize{38.223347235117004}{ identify}
\colorize{19.988494381278553}{ how}
\colorize{18.360731654712357}{ best}
\colorize{8.796509619470156}{\sdelete{ we}}
\colorize{0.27474613528257313}{\sdelete{ can}}
\colorize{60.65593629850552}{ shape}
\colorize{16.592888262379976}{\sdelete{ and}}
\colorize{22.369864268054364}{ develop}
\colorize{18.621241217306384}{ our services}
\colorize{6.6037275851597705}{\sdelete{ to}}
\colorize{14.714843680938586}{\sdelete{ support}}
\colorize{11.438700365569467}{\sdelete{ students}}
\colorize{48.824127267047174}{ most}
\colorize{3.5693015286695564}{\sdelete{ effectively}}
\colorize{10.98861324159768}{\sdelete{.}}
\colorize{34.986746476262844}{"}
\colorize{23.955086686381975}{We}
\colorize{46.523122030203126}{ work}
\colorize{25.060148569796276}{ closely}
\colorize{0.5399030932665699}{\sdelete{ with}}
\colorize{33.5579764269856}{ the three other universities}
\colorize{8.169351190774279}{\sdelete{ in}}
\colorize{41.25799161676557}{ Edinburgh and NHS mental health colleagues}
\colorize{6.835714066087542}{\sdelete{ to}}
\colorize{24.276447961154517}{ improve}
\colorize{56.53485851874477}{ pathways}
\colorize{19.910675208281855}{ into}
\colorize{12.27605012987393}{\sdelete{ specialist mental health services}}
\colorize{9.654175046660422}{\sdelete{ for}}
\colorize{11.503987519424538}{\sdelete{ students}}
\colorize{8.567184832923115}{\sdelete{.}}
\colorize{8.567184832923115}{\sdelete{"}}
\colorize{42.69109757625548}{ The Department}
\colorize{10.876024728604678}{\sdelete{ for}}
\colorize{5.269966371156043}{\sdelete{ Education}}
\colorize{56.034915971204}{ previously}
\colorize{42.036609799397816}{ responded}
\colorize{3.0206786132605106}{\sdelete{ to}}
\colorize{15.674527487000598}{\sdelete{ the petition}}
\colorize{10.08647317269807}{\sdelete{ by}}
\colorize{5.003436544501324}{\sdelete{ saying}}
\colorize{9.717945791849372}{\sdelete{ that}}
\colorize{24.06249548419218}{ higher education providers}
\colorize{28.79918153440236}{ already}
\colorize{11.138623199597196}{\sdelete{ had}}
\colorize{20.329418311827972}{ a "general duty}
\colorize{15.311282489620028}{\sdelete{ of}}
\colorize{0.7820664009994588}{\sdelete{ care}}
\colorize{53.53650137664334}{ not}
\colorize{0.6431957134176429}{\sdelete{ to}}
\colorize{25.09831211758088}{ cause}
\colorize{6.358358493612997}{\sdelete{ harm}}
\colorize{10.628063644519587}{\sdelete{ to}}
\colorize{5.635516538709007}{\sdelete{ their students}}
\colorize{33.02077810154837}{ through}
\colorize{11.965719526494334}{\sdelete{ their own actions}}
\colorize{6.272166869842683}{\sdelete{"}}
\colorize{6.272166869842683}{\sdelete{.}}
\colorize{42.93148019513329}{A spokesperson}
\colorize{31.7258073225362}{ said}
\colorize{6.60552065432156}{\sdelete{:}}
\colorize{5.081175281285627}{\sdelete{ "}}
\colorize{9.808998296159858}{\sdelete{We}}
\colorize{48.74893658344672}{ acknowledge}
\colorize{21.639835366197577}{ the profound and lasting impact}
\colorize{22.317470523796985}{ a young person's suicide}
\colorize{10.90149881982125}{\sdelete{ has}}
\colorize{32.72540330525288}{ upon}
\colorize{5.648311514133961}{\sdelete{ their family}}
\colorize{4.287168961836498}{\sdelete{ and}}
\colorize{0.6775638742264836}{\sdelete{ friends}}
\colorize{10.852169380381389}{\sdelete{,}}
\colorize{3.397124237907651}{\sdelete{ and}}
\colorize{42.57269854879123}{ know}
\colorize{100.0}{ among}
\colorize{33.63706046657852}{ the petitioners}
\colorize{14.670292467390036}{\sdelete{ there}}
\colorize{3.9330521148762747}{\sdelete{ are}}
\colorize{22.52820552840923}{ those}
\colorize{2.5228732787831594}{\sdelete{ who}}
\colorize{3.604486145659227}{\sdelete{ have}}
\colorize{28.381864867611295}{ personal experience}
\colorize{1.4940012341272249}{\sdelete{ of}}
\colorize{23.113100080362297}{ these devastating, tragic events}
\colorize{1.8382137641549408}{\sdelete{.}}
\colorize{43.02962233933269}{"[However] we}
\colorize{60.875017633591085}{...}
\colorize{87.03057675034465}{feel}
\colorize{79.64203117303423}{ further legislation}
\colorize{22.96226910049658}{ to}
\colorize{41.467906455875806}{ create}
\colorize{21.559345792053378}{ a statutory duty}
\colorize{8.808132795286644}{\sdelete{ of}}
\colorize{1.8920077711952383}{\sdelete{ care}}
\colorize{22.221792937273595}{,}
\colorize{36.22144018970287}{ where}
\colorize{11.797676637966378}{\sdelete{ such a duty}}
\colorize{14.071205131821706}{\sdelete{ already}}
\colorize{0.5358169837704625}{\sdelete{ exists}}
\colorize{8.441081032498134}{\sdelete{,}}
\colorize{15.636181812738634}{\sdelete{ would}}
\colorize{6.121243917894965}{\sdelete{ be}}
\colorize{16.882045950123466}{\sdelete{ a disproportionate response}}
\colorize{14.955219501382071}{\sdelete{.}}
\colorize{14.955219501382071}{\sdelete{"}}
\colorize{36.19795338818229}{ In}
\colorize{27.74326123895323}{ a response}
\colorize{3.1764067701494287}{\sdelete{ to}}
\colorize{22.187656746200062}{ the petition}
\colorize{2.521160288021435}{\sdelete{,}}
\colorize{16.257332500805465}{\sdelete{ the Scottish government}}
\colorize{8.820126664679844}{\sdelete{ said}}
\colorize{9.546083519065204}{\sdelete{:}}
\colorize{58.29368790279095}{ ''}
\colorize{12.99342500752894}{\sdelete{We}}
\colorize{13.486614205101704}{\sdelete{ are}}
\colorize{34.13372688440078}{ determined}
\colorize{1.4503204348592835}{\sdelete{ to}}
\colorize{28.754920148843325}{ support}
\colorize{16.742686872561624}{\sdelete{ the mental health}}
\colorize{5.958350783730476}{\sdelete{ of}}
\colorize{25.459257812911307}{ all students}
\colorize{17.852771964044386}{.}
\colorize{52.326754339041436}{Over}
\colorize{11.541352501644948}{\sdelete{ the last three years}}
\colorize{22.737978017846423}{ we}
\colorize{3.6138637628379517}{\sdelete{ have}}
\colorize{35.00484605530179}{ invested}
\colorize{16.680782858166694}{\sdelete{ £11.5m}}
\colorize{16.24061987393065}{\sdelete{ to}}
\colorize{39.22871687763466}{ introduce}
\colorize{33.72222319025168}{ additional counsellors}
\colorize{18.850102279070203}{ in}
\colorize{50.82649803724453}{ colleges}
\colorize{8.915070020296465}{\sdelete{ and}}
\colorize{9.738404069785696}{\sdelete{ universities}}
\colorize{53.244297513240525}{''}
\colorize{53.244297513240525}{.}
\colorize{34.57413482695981}{The BBC}
\colorize{16.22462924078083}{\sdelete{ is}}
\colorize{31.6017824117644}{ not}
\colorize{24.915223125044893}{ responsible}
\colorize{0.15348412968793812}{\sdelete{ for}}
\colorize{2.728974220092952}{\sdelete{ the content}}
\colorize{0.5036274349119314}{\sdelete{ of}}
\colorize{4.804561963172785}{\sdelete{ external sites}}
\colorize{2.250968681515996}{\sdelete{.}}